\documentclass{article}

\usepackage[main, final]{neurips_2026}

\usepackage[utf8]{inputenc} 
\usepackage[T1]{fontenc}    
\usepackage{hyperref}       
\usepackage{url}            
\usepackage{booktabs}       
\usepackage{amsfonts}       
\usepackage{nicefrac}       
\usepackage{microtype}      
\usepackage{xcolor}         

\usepackage{amsmath}
\usepackage{amssymb}
\usepackage{mathtools}
\usepackage{amsthm}
\usepackage{multirow}
\usepackage{microtype}
\usepackage{graphicx}
\usepackage{subcaption}
\usepackage{booktabs} 
\usepackage[dvipsnames,table]{xcolor}
\usepackage{array,multirow}
\usepackage{adjustbox}
\usepackage{subcaption,enumitem}
\usepackage{fontawesome5}
\usepackage{wrapfig}
\usepackage{etoolbox}
\usepackage{tocloft}     
\usepackage{titletoc}    
\usepackage{etoc}

\theoremstyle{plain}

\theoremstyle{definition}

\theoremstyle{remark}

\title{Unlocking Fine-Grained Perception in CLIP via Structurally-Aware Latent Masked Modeling}

\author{%
    Juntong Li \footnotemark[1]\textsuperscript{\hspace{0.6em}},
    Lingwei Dang 
    \thanks{Equal contributions. Email: qlzjftm@gmail.com, levondang@163.com.}\hspace{0.2em}
    \thanks{Project Leader.}\textsuperscript{\hspace{0.6em}},
    Haomin Wu\textsuperscript{\hspace{0.6em}},\\
    \textbf{Ziyan Qiu}\textsuperscript{\hspace{0.6em}}\textbf{,}
    \textbf{Qingxin Xiao}\textsuperscript{\hspace{0.6em}}\textbf{,}
    \textbf{Qingyao Wu}\thanks{Corresponding Author. Email: qyw@scut.edu.cn.}\textsuperscript{\hspace{0.6em}}\\
    School of Software Engineering, South China University of Technology
}

\begin{document}

\maketitle

\begin{abstract}
Vision-Language Models (VLMs) such as CLIP excel in global semantic alignment but often lack fine-grained perceptual capabilities. This hinders dense prediction tasks and bottlenecks the visual potential of Multimodal Large Language Models (MLLMs). Existing research has attempted to enhance CLIP's visual representations by incorporating geometric priors from vision-centric models. However, these strategies often struggle to achieve deep alignment for both local spatial structures and global semantics, potentially even distorting the original image-text space. To address these limitations, we propose SALM, an unsupervised embedding alignment framework based on structurally-aware latent mask modeling. SALM effectively synergizes local and global alignment via a dual-path design combining explicit and implicit mechanisms, without requiring external textual supervision. First, we introduce a dual-matrix alignment strategy that explicitly calibrates intra-sample spatial correlations and activation intensities, thereby effectively injecting local geometric priors. Based on this, we further design a latent mask modeling mechanism to guide CLIP to restore the missing semantic details of the target model, thereby implicitly aggregating fine-grained structures into the global semantic space. Furthermore, driven by the empirical observations that CLIP's shallow features inherently possess strong spatial observational capabilities, we naturally extend SALM to a highly efficient self-distillation paradigm, SALM-Self. This unlocks CLIP's intrinsic fine-grained potential without relying on any external models. Extensive experiments demonstrate that SALM not only significantly improves performance in dense prediction tasks but also boosts CLIP's zero-shot accuracy, effectively enhancing the fine-grained understanding capabilities of MLLMs. Project page at \href{https://qzfm.github.io/salm_project_page/}{https://qzfm.github.io/salm\_project\_page/}.
\end{abstract}
\section{Introduction}

\begin{figure}
    \centering
    \includegraphics[width=\textwidth]{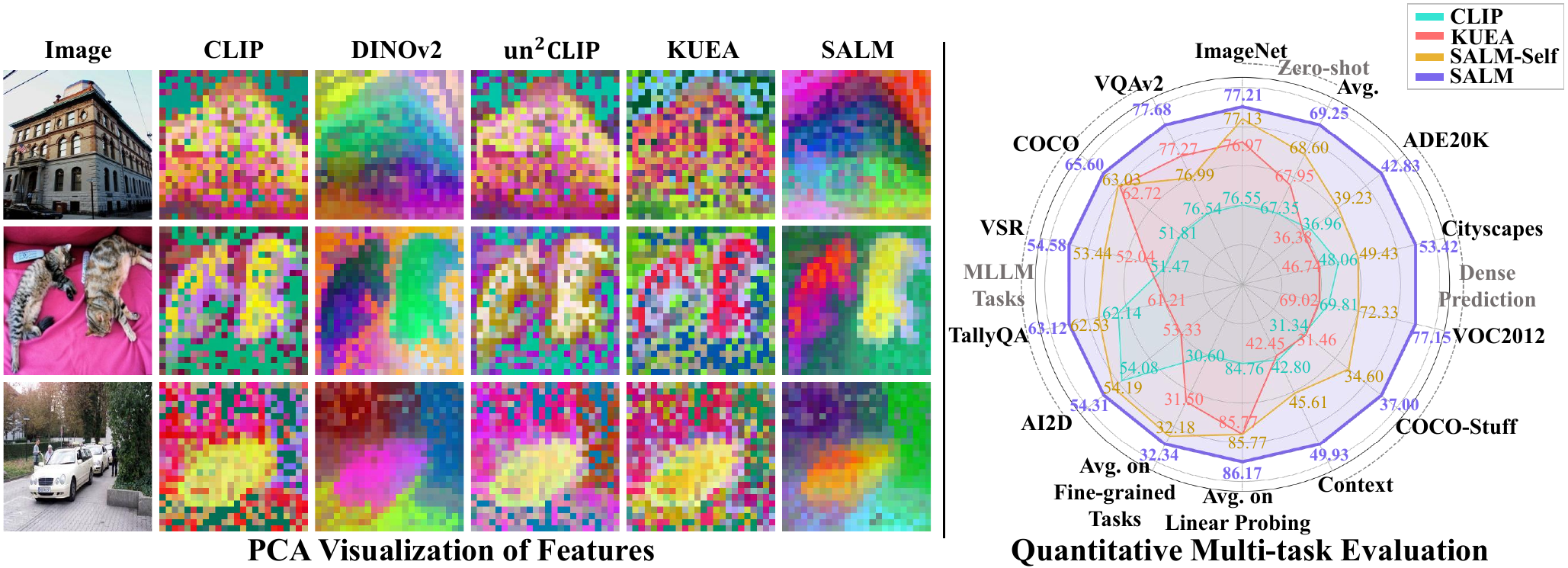}
    \caption{Comparison of feature visualizations and performance. Left: PCA visualization of features. In contrast to the fragmented and noisy outputs from CLIP and others, our method produces spatially coherent features with distinct semantic layouts. Right: The radar chart validates the superior performance of our method across multiple quantitative metrics.}
    \label{fig:teaser}
\end{figure}

Vision-Language Models (VLMs) have demonstrated remarkable generalization capabilities in multimodal understanding, content generation, and other downstream tasks. Among them, CLIP~\cite{radford2021learning} and its variants~\cite{zhai2023sigmoid, EVA-CLIP} rely on strong cross-modal alignment to maintain robust visual features even in zero-shot scenarios. However, constrained by contrastive learning's primary focus on global semantic consistency, CLIP often struggles to capture fine-grained information such as color, counting, and local geometric structure~\cite{sun2024alpha,lewis2024does}. This deficiency in fine-grained perception not only limits performance in tasks such as dense prediction~\cite{yan2025vision,xie2025region} but also constitutes a bottleneck for Multimodal Large Language Models (MLLMs)~\cite{liu2024improved, li2023blip, Qwen3-VL, zhu2025internvl3} aiming for refined visual understanding~\cite{guo2024llava,lou2025llava}.

Self-supervised vision models~\cite{darcet2024vision, simeoni2025dinov3, he2022masked} possess inherent advantages in comprehending local structures and spatial relationships. Effectively leveraging these priors to mitigate CLIP's limitations remains a challenge. Some approaches~\cite{jiang2023clip, huang2025ladi, zhou2025hints, wu2025when} employ multi-experts as enhanced visual encoders; however, they often lack the capacity to fundamentally enhance fine-grained perception at the perceptual level and introduce doubled computational overhead. Other methods attempt to distill diverse encoders into a single aggregated model~\cite{Ranzinger_2024_CVPR, heinrich2025radiov2, ranzinger2026c,zhu2026eupe} or fine-tune CLIP via generative reconstruction~\cite{wu2025vilau, ma2025unitok, wang2025visual, wang2025diffusion, li2025unclip, han2026dcr}. Nevertheless, these approaches either incur immense training costs or disrupt CLIP's original embedding distribution, severely compromising zero-shot capabilities or downstream compatibility, thereby limiting their practical applicability.

Recent works have explored kernel-based methods~\cite{jalali2025spec, gong2025kernel} to achieve efficient embedding alignment. Although kernel-based alignment can elegantly preserve the structure of the original space, it struggles to guarantee the spatial consistency of local dense features (\textit{e.g.}, KUEA~\cite{gong2025kernel} has no improvement over vanilla CLIP in segmentation tasks). This is because they primarily focus on the global alignment of embedding clusters between samples, neglecting the spatial positioning and intrinsic correlations of local features within samples, which restricts CLIP's fine-grained local perception. Furthermore, their performance is highly sensitive to the choice of the kernel function~\cite{gong2025kernel}, making it difficult to adaptively capture complex nonlinear correspondences between heterogeneous feature spaces.

Our key insight is that simultaneously achieving adaptive alignment of local spatial structures and global semantics can comprehensively address CLIP's fine-grained perception limitations, while avoiding disruption to the integrity of the original embedding space, preserving its zero-shot capability. We propose \textbf{SALM}, a novel framework for unsupervised embedding alignment via structurally-aware latent masked modeling. Despite requiring zero text supervision, SALM achieves adaptive alignment of local spatial consistency and fine-grained global semantics between CLIP and vision-centric models without breaking pre-trained image-text alignment.

We first introduce a Dual-Matrix Alignment (DMA) strategy to explicitly constrain the local spatial geometry at the manifold level. Specifically, we model the pairwise relationships between patches within the feature map by decoupling them into angular and magnitude components, constructing two key matrices: the Spatial Relation Matrix and the Energy Difference Matrix. By aligning these matrices, we effectively capture local semantic correlations and calibrate relative activation magnitudes, respectively. This mechanism explicitly injects fine-grained geometric priors from the target model into CLIP's space, enhancing local detail expression while maximizing the preservation of the original pre-trained distribution.

Building on explicit spatial constraints, we further aim to aggregate structural information into global semantics. This objective demands that the representation implicitly and intrinsically encodes and aligns with the detailed semantics of vision-centric models. In other words, these features must possess the capability to restore the fine-grained latents of these models. Inspired by masked modeling paradigm~\cite{he2022masked,zhang2022mask,lee2025latent,chen2025masked,naiman2025lv,wu2025harmon}, we propose an implicit alignment method based on latent masked modeling to bridge these distinct encoders. Specifically, we freeze the vision-centric encoder and apply high-ratio random masking to its dense features. Utilizing a lightweight Cross-Guided Adapter (CGA), we reconstruct the latents conditioned on both the visible context and the CLIP features. The unmasked patches serve as anchors, guiding the encoder to adaptively fill the missing fine-grained semantic gaps, effectively establishing a deep semantic alignment with the target feature space.

Furthermore, our empirical observations indicate that CLIP's shallow layers inherently capture rich local spatial details, which gradually diminish during deep feature aggregation. Inspired by this, we pose a natural question: can we unlock CLIP's intrinsic fine-grained potential directly within the SALM framework, without relying on any external vision-centric models? However, directly utilizing shallow features presents two major challenges: (1) they lack high-level semantics, meaning direct dual-matrix alignment would severely degrade the original deep semantics; and (2) they contain substantial low-level noise, making feature reconstruction difficult to optimize. To overcome these challenges, we introduce a small adjustment to our framework, extending it into a highly efficient self-distillation paradigm termed \textbf{SALM-Self}. Specifically, we discard the reconstruction loss and apply the DMA constraint on the reconstructed shallow features. This asymmetric design elegantly filters out low-level noise while strictly preserving the deep semantic integrity, enhancing CLIP's fine-grained perception and dense prediction capabilities with zero reliance on external models.

Experiments demonstrate that our method outperforms other alignment methods on dense prediction tasks, verifying that the model genuinely possesses fine-grained perception capabilities. Meanwhile, on multiple benchmarks, we achieve higher zero-shot accuracy without fine-tuning the text encoder. MLLMs integrated with our visual encoder exhibit superior performance over other methods.

In summary, the contributions of this paper are as follows:
\begin{itemize}
\item We propose SALM, a novel unsupervised embedding alignment framework bridging CLIP's global consistency and vision-centric models' fine-grained perception. It achieves simultaneous adaptive alignment of local geometry and global semantics.

\item We propose a dual-matrix alignment strategy to explicitly constrain the local spatial geometry at the manifold level. By aligning the spatial relation matrix and energy difference matrix, we effectively capture intrinsic spatial correlations, ensuring the transfer of spatial structures.

\item We design an implicit alignment based on latent masking modeling. Using the Cross-Guided Adapter, we guide the encoder to recover missing fine-grained details, establishing a deep and robust semantic alignment with the target feature space.

\item We extend the SALM to a highly efficient self-distillation paradigm, SALM-Self. It successfully unlocks CLIP's intrinsic fine-grained potential without relying on any external vision-centric models.

\end{itemize}

\section{Related Work}

\paragraph{Vision-Language Models.} As a milestone in Vision-Language Models (VLMs), CLIP~\cite{radford2021learning} and its variants (e.g., SigLIP~\cite{zhai2023sigmoid} and EVA-CLIP~\cite{EVA-CLIP}) demonstrate remarkable generalization capabilities in zero-shot tasks, attributed to robust cross-modal alignment established through large-scale contrastive learning. However, driven by a contrastive learning objective that overly prioritizes global consistency, CLIP often overlooks fine-grained details such as color, texture, and local geometric structures~\cite{sun2024alpha,lewis2024does}. This deficiency in fine-grained perception not only constrains performance in tasks like dense prediction~\cite{yan2025vision,xie2025region} but also serves as a primary bottleneck hindering MLLMs from achieving higher visual understanding precision~\cite{guo2024llava,lou2025llava}.

\paragraph{Enhancement of Visual Representations.} To address CLIP's limitations in fine-grained perception, researchers have attempted to incorporate prior knowledge from self-supervised vision models (e.g., DINOv2~\cite{oquab2024dinov, darcet2024vision}, MAE~\cite{he2022masked}, DINOv3~\cite{simeoni2025dinov3}). Some attempts predominantly adopted mixture-of-experts strategies~\cite{jiang2023clip, zhou2025hints,wu2025when}. While introducing additional information, this approach incurs multiplicative computational overheads and lacks deep feature fusion. Another category of methods explores distillation~\cite{Ranzinger_2024_CVPR,heinrich2025radiov2, ranzinger2026c, zhu2026eupe}, generative reconstruction~\cite{wu2025vilau, ma2025unitok, wang2025visual,wang2025diffusion,li2025unclip,han2026dcr} or architectural modifications~\cite{shi2026vision} to finetune CLIP. However, these approaches typically demand substantial data and computational resources and risk compromising its zero-shot capabilities. 
Alignment schemes based on kernel-based methods~\cite{jalali2025spec,gong2025kernel} have recently garnered attention for their efficiency, attempting to align feature matrices within a kernel space. However, kernel methods typically neglect spatial positional relationships and intrinsic correlations among local features within individual samples. Several works~\cite{wang2025declip,wysoczanska2024clip,lan2024proxyclip} have explored enhancing CLIP with vision-centric priors for open-vocabulary dense perception. In contrast, our goal is to develop a general CLIP enhancement framework that improves fine-grained representation while maintaining the global image-text alignment inherited from pre-trained CLIP.

\paragraph{Masked Modeling.} Masked Image Modeling (MIM)~\cite{he2022masked} facilitates the learning of local structures and spatial relationships by predicting masked image patches. Some works~\cite{lee2025latent,chen2025masked,naiman2025lv,wu2025harmon} have leveraged masked modeling to learn rich semantic representations and temporal dependencies. Recent theoretical studies~\cite{zhang2022mask} have provided insights into the effectiveness of MIM, suggesting that MIM essentially performs an implicit semantic alignment across samples. Inspired by this, we propose establishing a connection between frozen source and target features via latent masked modeling. 

\begin{wrapfigure}[8]{r}{0.5\textwidth}
\vspace{-7em}
    \centering
    \includegraphics[width=\linewidth]{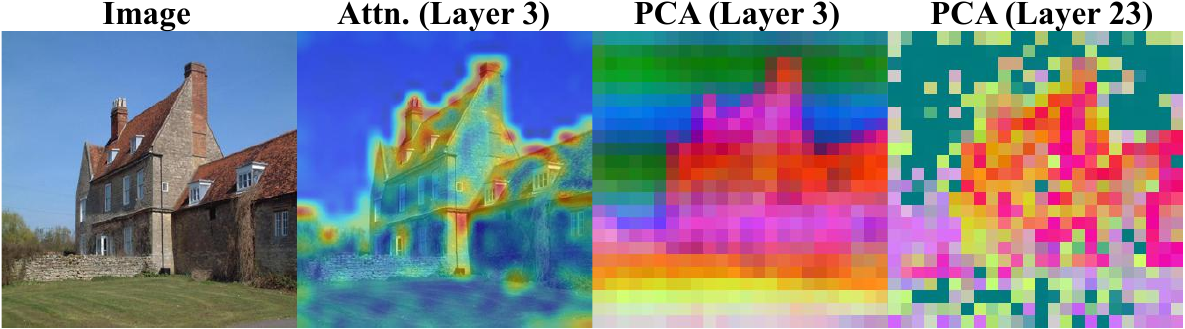}
    \caption{Visualization of CLIP’s shallow attention versus deep feature representation. CLIP demonstrates the capability to localize object textures (\textit{e.g.}, building edges) and detailed components (\textit{e.g.}, doors and windows) in its early layers, even though these details may not be effectively extracted in representations.}
    \label{fig:clip_attn}
\end{wrapfigure} 

\section{Method}
\label{method}

\subsection{Empirical Observations in CLIP}
\label{observations}

Motivated by CLIP's limited fine-grained performance, we visualized the attention maps within its shallow layers. As shown in Figure \ref{fig:clip_attn}, the results reveal that in the shallow layers, CLIP actually exhibits excellent observational capability, accurately covering various local parts of the target. However, it ultimately fails to effectively aggregate and preserve the texture and geometric details of these regions. This indicates that the bottleneck of CLIP's fine-grained perception lies in the final expressiveness of the representation rather than its initial observational capability.

This finding underscores the limitations of existing strategies: multi-expert strategies do not fundamentally improve the quality of CLIP's intrinsic fine-grained representation, while kernel-based methods tend to neglect the expressiveness of local features. And generative methods may distract the model with pixel-level redundancies.
Therefore, we do not intervene in the model's observation mechanism; instead, by introducing a vision-centric model to provide geometric priors, we drive the model at the feature manifold level to internalize fine-grained structural information through dual-matrix alignment and latent masked modeling. We additionally explore a self-distillation strategy, utilizing shallow features to unlock CLIP's fine-grained potential without relying on external experts.


\subsection{Overview}
\label{subsec:formulation}

Considering two distinct embedding spaces: a \textit{source space} defined by the visual encoder $f_\theta(\cdot)$ of a pre-trained VLM, and a \textit{reference space} characterized by a vision-centric model $g_\phi(\cdot)$ that encapsulates rich local structural priors. For a given image $x$, the extracted token sequences are denoted as $\mathbf{Z}^{\text{src}} = [ \mathbf{z}_{\text{cls}}^{\text{src}}; \mathbf{Z}_{\text{p}}^{\text{src}} ] \in \mathbb{R}^{(N+1) \times d}$ and $\mathbf{Z}^{\text{ref}} = [ \mathbf{z}_{\text{cls}}^{\text{ref}}; \mathbf{Z}_{\text{p}}^{\text{ref}} ] \in \mathbb{R}^{(N+1) \times d'}$, respectively. To address potential resolution mismatches, we spatially interpolate $\mathbf{Z}^{\text{src}}$ to the size of $\mathbf{Z}^{\text{ref}}$.

\begin{figure*}
    \centering
    \includegraphics[width=\textwidth]{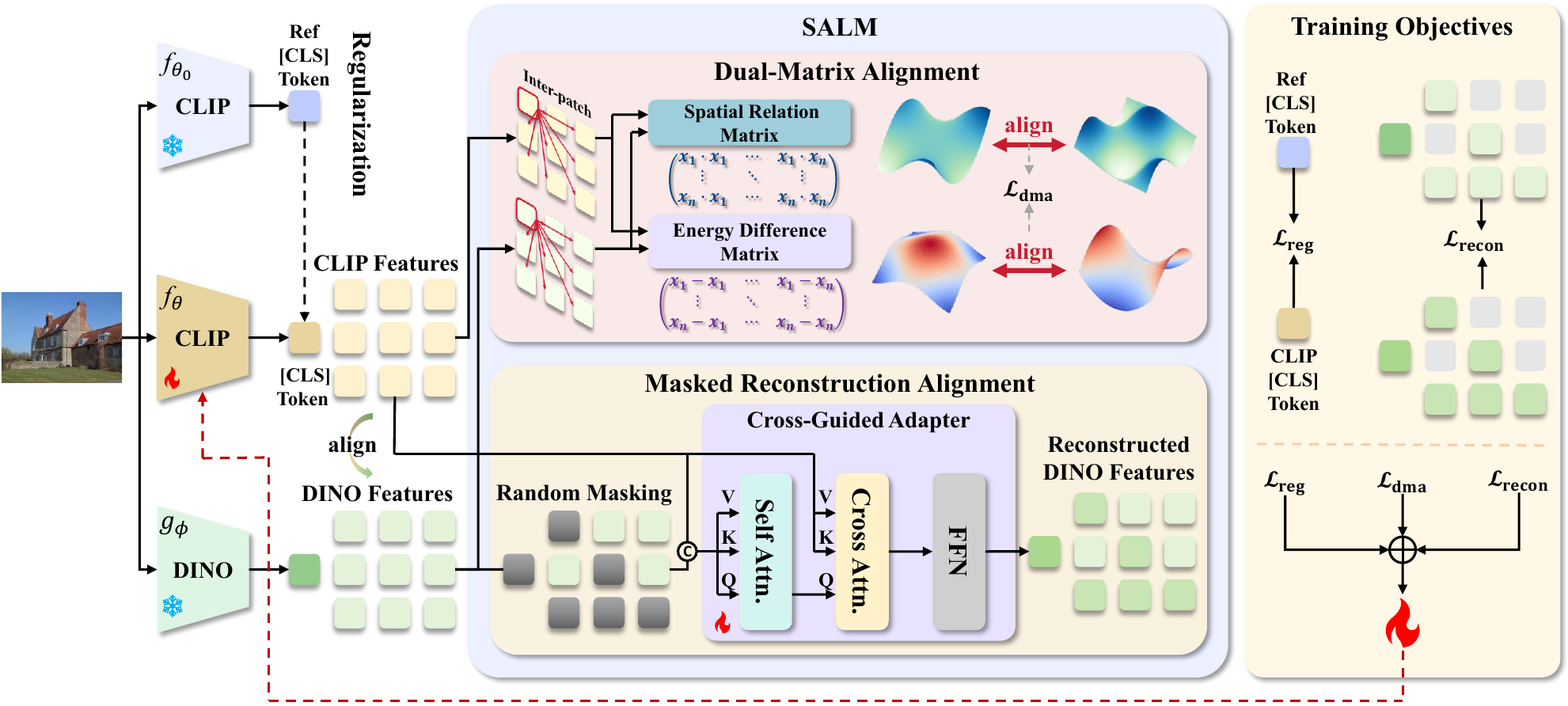}
    \caption{Overview of SALM. First, the Dual-Matrix Alignment strategy ensures relative consistency in spatial relationships and magnitude distributions (Sec. \ref{subsec:dma}). Subsequently, a latent masked modeling is employed to reconstruct the DINO features, thereby aggregating fine-grained information into global semantics (Sec. \ref{subsec:masked_modeling}). Finally, reference regularization is applied to the source encoder to ensure the stability of the adaptation process within the original multimodal latent space (Sec. \ref{subsec:Regularization}).}
    \label{fig:framework}
\end{figure*}

As illustrated in Figure~\ref{fig:framework}, our framework consists of a trainable source visual encoder $f_\theta$, a frozen reference encoder $g_\phi$, and a regularization encoder $f_{\theta_0}$. The Dual-Matrix Alignment (DMA) strategy (Sec. \ref{subsec:dma}) operates on the patch sequences of both encoders. Subsequently, we employ a latent space masked modeling approach (Sec. \ref{subsec:masked_modeling}) to further aggregate fine-grained semantics. We apply random masking to the reference features of the vision-centric model. A lightweight Cross-Guided Adapter (CGA) is tasked with reconstructing the reference tokens $\mathbf{Z}^{\text{ref}}$ by leveraging source space features and the visible reference features. Finally, we incorporate a regularization (Sec. \ref{subsec:Regularization})  to ensure the stability of the adaptation process within the original multimodal latent space. Additionally, we extend it into a self-distillation variant that unlocks fine-grained perception without external models (Sec. \ref{salm-self}).

\subsection{Dual-Matrix Alignment Strategy}
\label{subsec:dma}

Recent kernel-based methods~\cite{jalali2025spec, gong2025kernel} attempt to mitigate feature space heterogeneity; they are primarily confined to aligning inter-sample distribution clusters, neglecting the intrinsic geometric structure of intra-sample features. We argue that this oversight regarding local spatial relationships and relative saliency ranking constitutes the core bottleneck hindering CLIP's dense prediction capabilities. To address this, we propose the Dual-Matrix Alignment strategy, designed to explicitly capture and transfer second-order geometric statistics between patches at the manifold level, thereby injecting fine-grained structural priors without compromising the original semantic space.

Specifically, given the source feature sequence $\mathbf{Z}_{\text{p}}^{\text{src}}$ and reference feature sequence $\mathbf{Z}_{\text{p}}^{\text{ref}}$ for an image sample, we construct two key affinity matrices to describe the manifold. To capture semantic dependencies among local features, we define the Spatial Relation Matrix $\mathbf{K} \in \mathbb{R}^{N \times N}$: $\mathbf{K}_{ij}(\mathbf{Z}) = \frac{\mathbf{z}_i^\top \mathbf{z}_j}{\|\mathbf{z}_i\|_2 \|\mathbf{z}_j\|_2},$ where $\mathbf{K}_{ij}$ quantifies the semantic similarity between the $i$-th and $j$-th patches.


The magnitude of feature norms typically encodes the saliency or information entropy of local regions. To incorporate magnitude considerations while avoiding discrepancies caused by the varying magnitude scales of distinct models, we define the Energy Difference Matrix $\mathbf{D} \in \mathbb{R}^{N \times N}$, utilizing normalization to eliminate dimensional scale effects: $\mathbf{D}_{ij}(\mathbf{Z}) = \mathcal{Z}(\|\mathbf{z}_i\|_2 - \|\mathbf{z}_j\|_2),$ where $\mathcal{Z}$ denotes Z-score normalization within the matrix, and $\mathbf{D}_{ij}$ characterizes the relative saliency ranking.

The total objective of the DMA strategy is defined as the sum of the Frobenius norm distances of these two matrices: $ \mathcal{L}_{\text{dma}} = \| \mathbf{K}(\mathbf{Z}^{\text{src}}_\text{p}) - \mathbf{K}(\mathbf{Z}^{\text{ref}}_\text{p}) \|_F^2 + \| \mathbf{D}(\mathbf{Z}^{\text{src}}_\text{p}) - \mathbf{D}(\mathbf{Z}^{\text{ref}}_\text{p}) \|_F^2.$

\subsection{Latent Masked Modeling for Implicit Alignment}
\label{subsec:masked_modeling}

While explicit alignment rectifies source spatial geometry, second-order constraints fail to integrate fine details into global semantics. To address this, we adopt latent masked modeling to aggregate spatial structural information, leveraging source features as semantic anchors to reconstruct missing details and global semantics within reference features for deep semantic integration.


Specifically, given the reference sequence $\mathbf{Z}^{\text{ref}}$, we apply a high-ratio (\textit{\textit{e.g.}}, 75\%) random masking $M \in \{0,1\}^N$ to partition it into visible context $x_1 = \mathbf{Z}^{\text{ref}}[M]$ and masked target $x_2 = \mathbf{Z}^{\text{ref}}[1-M]$, replacing dropped patches with a learnable [MASK] token.

To facilitate reconstruction within the latent space and mitigate the risk of the decoder dominating representation learning, we design a lightweight Cross-Guided Adapter(CGA). The CGA recovers the target $x_2$ utilizing the complete source features $\mathbf{Z}^{\text{src}}$ and the sparse context $x_1$. It first concatenates the full source features with the visible reference context to complement the spatial positional basis. Subsequently, in the cross-attention layer, the fused features serve as the query, while the source features are explicitly designated as key and value. The process of CGA is defined as follows:
\begin{equation}
\begin{aligned}
    \mathbf{H}_{\text{mid}} &= \text{SelfAttn}\big([\mathbf{Z}^{\text{src}} ; x_1]\big) + [\mathbf{Z}^{\text{src}} ; x_1], \\
    \mathbf{H}_{\text{out}} &= \text{CrossAttn}\big(\mathbf{H}_{\text{mid}}, \mathbf{Z}^{\text{src}}, \mathbf{Z}^{\text{src}}\big) + \mathbf{H}_{\text{mid}}, \\
    \hat{x}_2 &= \mathcal{P}_{\text{target}}\big(\text{FFN}(\mathbf{H}_{\text{out}})\big),
\end{aligned}
\end{equation}
where $\mathcal{P}_{\text{target}}$ represents the slicing operator extracting features corresponding to the target regions. The optimization objective is to minimize the reconstruction error: $\mathcal{L}_{\text{recon}} =  \| \hat{x}_2 - x_2 \|_2^2.$

\begin{wrapfigure}[11]{r}{0.5\textwidth}
\vspace{-2.5em}
    \centering
    \includegraphics[width=0.9\linewidth]{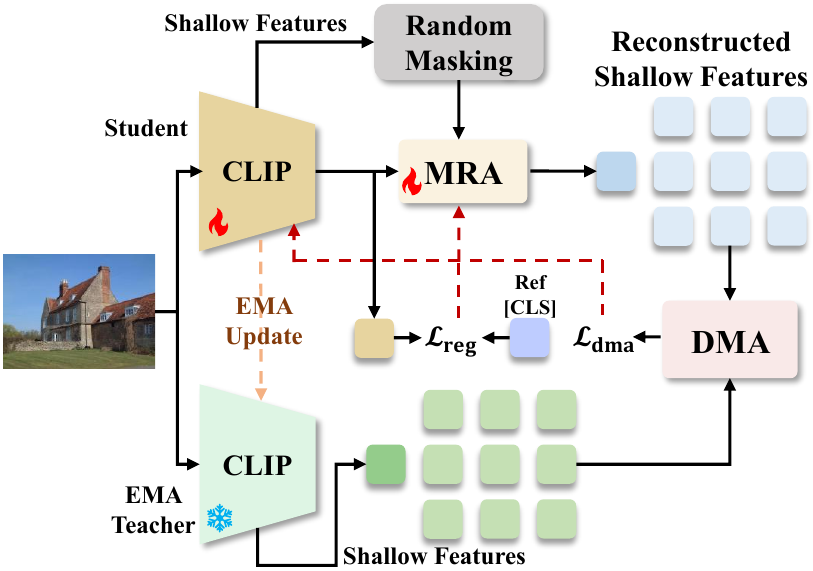}
    
    \caption{Overview of SALM-Self. }
    \label{fig:self}
\end{wrapfigure} 

\subsection{Regularization for Zero-shot Preservation}
\label{subsec:Regularization}

Following ~\cite{schlarmann2024robustclip,gong2025kernel}, to maintain the pre-trained image-text alignment, we introduce a regularization term $\mathcal{L}_{\text{reg}} = \| f_\theta(x) - f_{\theta_0}(x) \|_2^2$, where $f_{\theta_0}$ denotes the frozen initial source encoder. This ensures that the language-image alignment is effectively preserved even without incorporating any textual data during the fine-tuning phase.

The final objective is a balanced combination of the aforementioned constraints:$\mathcal{L} = \alpha \mathcal{L}_{\text{reg}} + \beta \mathcal{L}_{\text{recon}} + \gamma \mathcal{L}_{\text{dma}} + \lambda \mathcal{L}_{\text{uni}}, $where $\mathcal{L}_{\text{uni}}$ denotes the uniformity loss~\cite{zhang2022mask}. $\alpha, \beta, \gamma,$ and $\lambda$ are hyperparameters.

\subsection{SALM-Self: Enhancing CLIP via Intra-Model Alignment}
\label{salm-self}
As discussed in Section \ref{observations}, while CLIP's shallow layers inherently capture local geometric information, these details gradually diminish during deep feature aggregation. However, directly aligning or reconstructing these shallow features is impractical due to their lack of high-level semantics and the presence of substantial low-level noise. With minor adjustments to SALM, we adapt it into a CLIP-based self-distillation architecture, SALM-Self, as shown in Figure \ref{fig:self}. Specifically, we set CLIP's shallow features as the reference features $\mathbf{Z}^{\text{ref}}$ and discard the reconstruction loss $\mathcal{L}_{\text{recon}}$. In its place, we perform dual-matrix alignment between the reconstructed shallow features and those of the teacher model, whose parameters are dynamically updated via an exponential moving average (EMA). The total objective loss is: $\mathcal{L} = \alpha \mathcal{L}_{\text{reg}} + \gamma \mathcal{L}_{\text{dma}} + \lambda \mathcal{L}_{\text{uni}}$.


\section{Experiments}
\label{experiments}

\subsection{Experimental Setup}

\paragraph{Implementation Details.}

We adopt OpenAI CLIP ViT-L-14-336~\cite{radford2021learning} and DINOv2 ViT-L/14 with registers~\cite{darcet2024vision} as the default VLM and vision-centric backbone, respectively. Notably, our framework is designed to be model-agnostic and resolution-flexible, capable of aligning arbitrary VLMs with various vision models. 
Our default training pipeline adopts a three-stage curriculum strategy to progressively optimize structural alignment and fine-grained semantic reconstruction. For SALM-Self, we use the 6th CLIP layer by default as the self-reference feature, which provides a balance between local structural details and semantic consistency.
Extensive experiments demonstrating this generalization across different architectures and more details are provided in the \textbf{Appendix}.

We train exclusively on the ImageNet-1K~\cite{deng2009imagenet} training set. In contrast to the massive datasets required for CLIP's pre-training or the distillation processes of aggregation models, our data regime is exceptionally lightweight and purely image-based, independent of text. For inference, the auxiliary CGA is removed, leaving only the finetuned CLIP for downstream tasks to ensure zero extra cost.

\paragraph{Compared Methods.}
We benchmark our method against several SOTA approaches for improving CLIP, including: (1) RADIOv2.5~\cite{heinrich2025radiov2}, an aggregation method that distills diverse vision models into a single one. For a fair comparison, we reproduced it on ImageNet-1K following its protocol: using CLIP and DINOv2 as teachers and a student model that matches the CLIP vision encoder in both architecture and initialization; (2) un$^2$CLIP~\cite{li2025unclip}, a generative-based approach that employs unCLIP~\cite{ramesh2022hierarchical} to inject fine-grained semantics into CLIP; and (3) KUEA~\cite{gong2025kernel}, a kernel-based alignment method which is also trained on ImageNet-1K that employs kernel matrix alignment to mitigate the disruption of the original space. All compared methods maintain identical model sizes.

\begin{table*}[!t]
    \centering
    \caption{Zero-shot object recognition performance on various benchmarks. We report Top-1 accuracy (\%) on ImageNet-1K and 11 other datasets. The best results are highlighted in \textbf{bold}. $^{\dagger}$ indicates the model reproduced by us on ImageNet-1K.}
    \label{tab:zs_cls}
    \resizebox{\textwidth}{!}{
        \begin{tabular}{l | c | cccccccccccc}
            \toprule
            \multirow{7}{*}{\textbf{Method}} & 
            \multirow{7}{*}{\rotatebox{270}{ImageNet}} & 
            \multicolumn{12}{c}{\textbf{Zero-Shot}} \\
            
            \cmidrule(lr){3-14}
            
             & & 
            \rotatebox{270}{CIFAR-10} & \rotatebox{270}{CIFAR-100} & \rotatebox{270}{Pets} & 
            \rotatebox{270}{Caltech-101} & \rotatebox{270}{RESISC45} & \rotatebox{270}{PCam} & 
            \rotatebox{270}{DTD} & \rotatebox{270}{EuroSAT} & 
            \rotatebox{270}{ImageNet-O} & \rotatebox{270}{FER2013} & \rotatebox{270}{ImageNetV2} & 
            \rotatebox{270}{\textbf{Average}} \\ 
            \midrule
            
            CLIP & 76.55 & 94.92 & 74.33 & 93.68 & 83.43 & 63.78 & 60.72 & 55.69 & 61.50  & 32.75 & 49.11 & 70.89 & 67.35 \\

            RADIOv2.5$^\dagger$ & 75.02 & 90.97 & 64.94 & 86.70 & 83.78 & 37.35 & 50.02 & 43.62 & 26.69 & \textbf{61.05} & 34.68 & 67.82 & 58.87 \\
            
            un$^2$CLIP & 71.25 & 91.84 & 69.95 & 91.74 & \textbf{85.45} & 57.33 & 61.78 & 52.93 & \textbf{62.46} & 36.15 & \textbf{51.49} & 64.91 & 66.00 \\
            
            KUEA & 76.97 & \underline{95.86} & 76.95 & 93.79 & 83.43 & \underline{64.43} & 59.60 & \textbf{56.33} & 61.89 & 35.65 & 48.08 & \underline{71.40} & 67.95 \\

            {\cellcolor{blue!10}}{\textbf{SALM-Self (Ours)}} &{\cellcolor{blue!10}\textit{\underline{77.13}}} &{\cellcolor{blue!10}\textit{95.58}} &{\cellcolor{blue!10}\textit{\underline{77.37}}} &{\cellcolor{blue!10}\textit{\underline{93.87}}} &{\cellcolor{blue!10}\textit{83.79}} &{\cellcolor{blue!10}\textit{63.95}} &{\cellcolor{blue!10}\textit{\underline{64.74}}} &{\cellcolor{blue!10}\textit{55.96}} &{\cellcolor{blue!10}\textit{61.72}} &{\cellcolor{blue!10}\textit{36.80}} &{\cellcolor{blue!10}\textit{49.60}} &{\cellcolor{blue!10}\textit{71.27}} &{\cellcolor{blue!10}\textit{\underline{68.60}}}\\
            
            {\cellcolor{blue!10}}\textbf{SALM (Ours)} &{\cellcolor{blue!10}{\textbf{77.21}}} &{\cellcolor{blue!10}{\textbf{96.38}}} &{\cellcolor{blue!10}{\textbf{77.87}}} &{\cellcolor{blue!10}{\textbf{94.09}}} &{\cellcolor{blue!10}{\underline{84.04}}} &{\cellcolor{blue!10}{\textbf{64.63}}} &{\cellcolor{blue!10}{\textbf{67.54}}} &{\cellcolor{blue!10}{\underline{56.27}}} &{\cellcolor{blue!10}{\underline{62.19}}} &{\cellcolor{blue!10}{\underline{37.05}}} &{\cellcolor{blue!10}{\underline{50.11}}} &{\cellcolor{blue!10}{\textbf{71.53}}} &{\cellcolor{blue!10}{\textbf{69.25}}} \\
            
            \bottomrule
        \end{tabular}
    }
\end{table*}

\begin{table}[t]
\centering
\caption{Quantitative results on zero-shot fine-grained understanding and dense prediction. We report Top-1 accuracy (\%) for zero-shot fine-grained understanding tasks, alongside mean IoU (mIoU) for semantic segmentation. $^{\dagger}$ indicates the model reproduced by us on ImageNet-1K.}

\label{tab:fine_seg}
\resizebox{\linewidth}{!}{
\begin{tabular}{l|ccc|ccccc}
\toprule
\multirow{2.5}{*}{\textbf{Method}} & \multicolumn{3}{c|}{\textbf{ZS. Fine-grained Tasks}} & \multicolumn{5}{c}{\textbf{Linear Probing Segmentation}} \\
 & SVHN  &\adjustbox{stack=cc}{CLEVR\\Distance}&\adjustbox{stack=cc}{CLEVR\\Counts} &ADE20K  & Cityscapes & VOC2012 & COCO-Stuff & Context\\
\midrule
CLIP      & 55.97 & 15.81 & 20.01 &  36.96 &  48.06 & 69.81 & 31.34 & 42.80\\
RADIOv2.5$^\dagger$  & 41.25  & \textbf{19.45} & 15.60 & \underline{42.35} & \underline{51.82} & \underline{75.31} & \underline{35.11} & \underline{48.73}\\
un$^2$CLIP & 54.77 & 15.93 & 22.73 & 37.46 & 48.11 & 71.64 & 32.74 & 44.44\\
KUEA       & 57.74 & 15.95 & 20.81 & 36.38 & 46.74 & 69.02 & 31.46 & 42.45\\

{\cellcolor{blue!10}}\textbf{{SALM-Self (Ours)}} & {\cellcolor{blue!10}}\textit{\underline{57.92}} & {\cellcolor{blue!10}}\textit{15.83} & {\cellcolor{blue!10}}\underline{22.78} & {\cellcolor{blue!10}}\textit{39.23} & {\cellcolor{blue!10}}\textit{49.43} & {\cellcolor{blue!10}}\textit{72.33} & {\cellcolor{blue!10}}\textit{34.60} & {\cellcolor{blue!10}}\textit{45.61}\\

{\cellcolor{blue!10}}\textbf{SALM (Ours)} & 
{\cellcolor{blue!10}}\textbf{57.99} &  
{\cellcolor{blue!10}}\underline{16.21} & 
{\cellcolor{blue!10}}\textbf{22.81} & 
{\cellcolor{blue!10}}\textbf{42.83} &
{\cellcolor{blue!10}}\textbf{53.42} &
{\cellcolor{blue!10}}\textbf{77.15} &
{\cellcolor{blue!10}}\textbf{37.00} &
{\cellcolor{blue!10}}\textbf{49.93}\\
\bottomrule
\end{tabular}
}
\vspace{-3mm} 
\end{table}

\paragraph{Evaluated Tasks and Benchmarks.}


We conduct evaluations across multiple benchmarks:
\begin{itemize}
\item Object Recognition: (1) General Objects (ImageNet~\cite{deng2009imagenet}, CIFAR-10/100~\cite{krizhevsky2009learning}, Caltech-101~\cite{FeiFei2004LearningGV}); (2) Fine-grained Domains (Pets~\cite{parkhi12a}, DTD~\cite{cimpoi14describing}, EuroSAT~\cite{helber2018introducing}, FER2013~\cite{goodfellow2013challenges}); (3) Specialized Domains (RESISC45~\cite{cheng2017remote}, PCam~\cite{Veeling2018-qh}, GTSRB~\cite{stallkamp2012man}); and (4) Distribution Shifts (ImageNet-O~\cite{hendrycks2021nae}, ImageNetV2~\cite{recht2019imagenet}).


\item Fine-grained Understanding: (1) Object Counting and Spatial Reasoning (CLEVR~\cite{johnson2017clevr}), and (2) Complex Number Recognition (SVHN~\cite{netzer2011reading}).

\item Dense Prediction: We perform linear probing for semantic segmentation on ADE20K~\cite{zhou2019semantic}, Cityscapes~\cite{cordts2016cityscapes}, PASCAL VOC 2012~\cite{everingham2015pascal}, COCO-Stuff~\cite{caesar2018coco}, and PASCAL Context~\cite{mottaghi2014role}.

\item MLLM Integration: We perform our experiments using LLaVA-1.5-7B~\cite{liu2024improved}. For fair comparison, following~\cite{gong2025kernel}, all methods skip the feature alignment step and utilize the original SFT dataset for visual instruction tuning via LoRA~\cite{hu2022lora}. The evaluation benchmarks include: (1) Open-ended VQA: VQA v2~\cite{goyal2017making}; (2) Diagram Understanding: AI2D~\cite{kembhavi2016diagram}; (3) Spatial Awareness: RefCOCO, RefCOCO+, RefCOCOg~\cite{kazemzadeh2014referitgame,yu2016modeling}, and VSR~\cite{liu2023visual}; (4) Counting Ability: TallyQA~\cite{acharya2019tallyqa}; and (5) Hallucination Evaluation: POPE~\cite{li2023evaluating}.
\end{itemize}

\subsection{Evaluation on Vision-centric Tasks}

\paragraph{Zero-shot Object Recognition.}

As presented in Table \ref{tab:zs_cls}, we evaluate zero-shot classification performance on ImageNet-1K and 11 downstream datasets. Our method outperforms all baselines, achieving an average accuracy of 69.25\% across the 11 downstream tasks. 
First, the distillation strategy used by RADIOv2.5, particularly with limited training data, yields only marginal gains and frequently fails to retain zero-shot capabilities. This confirms that simple multi-model aggregation is often insufficient.
Second, compared to KUEA, we not only achieve higher accuracy on ImageNet-1K but also surpass it in average accuracy. This indicates that the kernel-based method can only achieve sub-optimal alignment. In contrast, our methods capture the fine-grained geometric structures of vision-centric models more effectively. Finally, we notice that un$^2$CLIP exhibits alignment degradation across multiple datasets. This suggests that its generative strategy disrupts the distribution stability of the original feature space. Conversely, SALM effectively enhances visual representation while preserving the integrity of the original distribution and CLIP's zero-shot generalization.

\begin{table*}[!t]
    \centering
    \caption{Linear probing classification performance on various benchmarks. We report Top-1 accuracy (\%) on 14 datasets. The best results are highlighted in \textbf{bold}. $^{\dagger}$ indicates the model reproduced by us on ImageNet-1K.}
    \label{tab:appendix_lp}
    \resizebox{\textwidth}{!}{
        \begin{tabular}{l | ccccccccccccccc}
            \toprule
            \multirow{4.5}{*}{\textbf{Method}} &
            \rotatebox{270}{ImageNet} &\rotatebox{270}{CIFAR-10} & \rotatebox{270}{CIFAR-100} & \rotatebox{270}{Pets} & \rotatebox{270}{SVHN} & 
            \rotatebox{270}{Caltech-101} & \rotatebox{270}{RESISC45} & \rotatebox{270}{PCam} & 
            \rotatebox{270}{DTD} & \rotatebox{270}{EuroSAT} & \rotatebox{270}{GTSRB} & 
            \rotatebox{270}{CLEVR D.} & \rotatebox{270}{FER2013} & \rotatebox{270}{CLEVR C.} & 
            \rotatebox{270}{\textbf{Average}} \\ 
            \midrule
            
            CLIP & 80.09 & 97.42 & 85.49 & 94.73 & 78.5 & 95.97 & 96.01 & 84.46 & 80.58 & 97.14 & 92.76 & 55.88 & 71.92 & 75.62 & 84.76 \\

            RADIOv2.5$^\dagger$& 80.43 &  97.30 & 86.36 & 94.71 & 80.56 & 96.31 & 96.00 & 84.48 & 81.34 & 97.05 & 91.14 & 56.08 & 72.11 & 76.86 & 85.05 \\

            un$^2$CLIP& 80.02  & 97.02 & 84.51 & 93.76 & \textbf{81.63} & \textbf{96.53} & 95.82 & 84.14 & 80.37 & \underline{97.70} & 92.5 & \textbf{60.59} & 72.21 & 77.43 & 85.30 \\
            
            KUEA & 80.72 & 98.13 & 87.83 & 94.93 & 81.48 & \underline{96.49} & \underline{96.25} & \underline{84.99} & \textbf{81.86} & 97.40 & \underline{93.63} & 57.07 & 72.12 & 77.83 & \underline{85.77} \\

            {\cellcolor{blue!10}}\textbf{{SALM-Self (Ours)}} &{\cellcolor{blue!10}{\textbf{\textit{81.57}}}}&{\cellcolor{blue!10}{\underline{\textit{98.16}}}} &{\cellcolor{blue!10}{\underline{\textit{88.23}}}}&{\cellcolor{blue!10}{\underline{\textit{95.11}}}} &{\cellcolor{blue!10}{\textit{81.40}}}  &{\cellcolor{blue!10}{\textit{96.42}}} &{\cellcolor{blue!10}{\textit{96.06}}} &{\cellcolor{blue!10}{\textit{84.59}}} &{\cellcolor{blue!10}{\underline{\textit{81.55}}}} &{\cellcolor{blue!10}{\textit{97.02}}} &{\cellcolor{blue!10}{\textit{93.39}}} &{\cellcolor{blue!10}{\textit{57.01}}} &{\cellcolor{blue!10}{\underline{\textit{72.39}}}} &{\cellcolor{blue!10}{\underline{\textit{77.85}}}} &{\cellcolor{blue!10}{\underline{\textit{85.77}}}}\\
            
            {\cellcolor{blue!10}}\textbf{SALM (Ours)} &{\cellcolor{blue!10}{\underline{80.94}}}&{\cellcolor{blue!10}{\textbf{98.17}}} &{\cellcolor{blue!10}{\textbf{88.28}}}&{\cellcolor{blue!10}{\textbf{95.20}}} &{\cellcolor{blue!10}{\underline{81.61}}}  &{\cellcolor{blue!10}{96.45}} &{\cellcolor{blue!10}{\textbf{96.90}}} &{\cellcolor{blue!10}{\textbf{85.49}}} &{\cellcolor{blue!10}{\textbf{81.86}}} &{\cellcolor{blue!10}{\textbf{97.73}}} &{\cellcolor{blue!10}{\textbf{93.66}}} &{\cellcolor{blue!10}{\underline{57.88}}} &{\cellcolor{blue!10}{\textbf{72.50}}} &{\cellcolor{blue!10}{\textbf{79.75}}} &{\cellcolor{blue!10}{\textbf{86.17}}} \\
            
            \bottomrule
        \end{tabular}
    }
\end{table*}

\begin{table*}[!t]
    \centering
    \caption{Performance evaluation on MLLM benchmarks. We compare the performance of different visual encoders when integrated into the LLaVA. ``+FT'' indicates that the model was fine-tuned using the same SFT data.}
    \label{tab:mllm}
    \resizebox{\textwidth}{!}{
        \begin{tabular}{l | ccccccccc}
            \toprule
            \textbf{Method} & 
            AI2D & POPE & TallyQA & VSR & 
            RefCOCO & RefCOCO+ & RefCOCOg & 
            VQA v2 & 
            \textbf{Average} \\ 
            \midrule
            
            LLaVA-7B & 54.08 & 86.89 & 62.14 & 51.47 & 54.86 & 49.55 & 51.04 & 76.54 & 60.82 \\
            + FT & 53.95 & 86.90 & 62.45 & 52.70 & 66.42 & 58.64 & 58.93 & 77.23 & 64.65 \\
            \midrule

            un$^2$CLIP & 52.92 & 86.93 & \underline{62.75} & 51.47 & 65.42 & 60.66 & 59.82 & \underline{77.40} & 64.68 \\
            
            KUEA & 53.33 & \textbf{87.30} & 61.21 & 52.04 & 66.94 & \underline{60.74} & 60.48 & 77.27 & 64.92 \\

            {\cellcolor{blue!10}}{\textbf{SALM-Self (Ours)}} & 
            {\cellcolor{blue!10}\textit{\underline{54.19}}} & {\cellcolor{blue!10}\textit{86.92}} & {\cellcolor{blue!10}\textit{62.53}} & 
            {\cellcolor{blue!10}\textit{\underline{53.44}}} & {\cellcolor{blue!10}\textit{\underline{66.99}}} & {\cellcolor{blue!10}\textit{\underline{60.74}}} & {\cellcolor{blue!10}\textit{\underline{61.37}}} & {\cellcolor{blue!10}\textit{76.99}} & 
            {\cellcolor{blue!10}\textit{\underline{65.40}}} \\
            
            {\cellcolor{blue!10}}\textbf{SALM (Ours)} &
            {\cellcolor{blue!10}{\textbf{54.31}}} & {\cellcolor{blue!10}{\underline{87.06}}} & {\cellcolor{blue!10}{\textbf{63.12}}} & 
            {\cellcolor{blue!10}{\textbf{54.58}}} & {\cellcolor{blue!10}{\textbf{69.63}}} & {\cellcolor{blue!10}{\textbf{63.68}}} & {\cellcolor{blue!10}{\textbf{63.48}}} & {\cellcolor{blue!10}{\textbf{77.68}}} & 
            {\cellcolor{blue!10}{\textbf{66.69}}} \\
            
            \bottomrule
        \end{tabular}
    }
\end{table*}

\paragraph{Fine-grained Understanding.}

We further evaluate fine-grained understanding capabilities in Table \ref{tab:fine_seg}. Our method achieves the best performance on the geometry-sensitive CLEVR benchmarks and the SVHN recognition task. Specifically, it reaches 22.81\% on CLEVR Counts, significantly surpassing all baselines. This directly demonstrates that the geometric constraints introduced via DMA successfully endow the model with the core capability to handle local spatial relationships and object counting, effectively bridging the perceptual gap inherent in CLIP. In contrast, un$^2$CLIP suffers from severe performance degradation on SVHN, further corroborating that its generative injection strategy disrupts the stability of the original feature distribution. Conversely, our approach maintains superior robustness while significantly enhancing structural understanding.

\paragraph{Dense Prediction.}
To validate the quality of local dense features, we conducted linear probing semantic segmentation experiments on multiple benchmark datasets (see Table \ref{tab:fine_seg}, right). For instance, it achieves substantial improvements on COCO-Stuff and VOC2012, reaching 37.00 and 77.15 mIoU, respectively. Notably, the kernel-based alignment method, KUEA, performs slightly worse than the original CLIP on this task. This aligns with our expectations: while kernel methods effectively align global semantic distributions, they neglect the intrinsic local structural information within samples. Furthermore, although other methods attempt to enhance representations through various strategies, none match our level of dense prediction capability.


\paragraph{Linear Probing.} To further validate the robustness and linear separability of the learned visual representations under frozen parameters, we evaluated the model's classification performance without reliance on textual priors. We followed the linear probing protocol, where the backbone is frozen and only a linear head is trained. As shown in Table \ref{tab:appendix_lp}, our method achieved an average accuracy of 86.17\%, outperforming existing methods designed for enhancing VLM visual representations. This further corroborates that our enhanced visual representations capture richer semantic structural information, yielding high-quality features that are both more discriminative and fine-grained.

\subsection{Multimodal Large Language Model Evaluation}

To evaluate the efficacy of our aligned encoder on MLLMs, we integrated the SALM-aligned encoder into the LLaVA framework, benchmarking it against the vanilla encoder and the kernel-based KUEA method. As reported in Table \ref{tab:mllm}, our approach delivers an SOTA average score of 66.69\%, surpassing both the fine-tuned LLaVA baseline and KUEA by a significant margin. It is worth noting that, enabled by the explicit constraints on local geometric information via our dual-matrix alignment strategy, our model demonstrates superior performance in tasks requiring spatial reasoning. Specifically, compared to KUEA, our method improves by 2.69\%, 2.94\%, and 3.00\% on RefCOCO, RefCOCO+, and RefCOCOg, respectively. Moreover, the performance boosts on VSR and TallyQA further confirm that SALM effectively addresses CLIP's limitations in fine-grained perception and local structure capture, successfully elevating overall multimodal understanding.


\begin{wraptable}[11]{r}{0.5\linewidth}
\raggedleft
\vspace{-4em}
    \caption{Ablation study on the effectiveness of proposed components. We analyze the impact of the DMA and CGA modules. ``+FT'' represents finetuning CLIP solely on ImageNet-1K with the text ``This is a photo of \{class name\}''.}
    \label{tab:ablation}
    
    \centering
    \resizebox{0.5\textwidth}{!}{
    
        \begin{tabular}{l|cccc|c}
        \toprule
        \multirow{2.5}{*}{\textbf{Method}} & \multicolumn{4}{c|}{\textbf{ZS. Cls.}} & \multicolumn{1}{c}{\textbf{LP. Seg.}} \\
         & RESISC & PCam & SVHN &\adjustbox{stack=cc}{CLEVR\\Distance} & ADE20K\\
        \midrule
        CLIP       & 63.78 & 60.72 & 55.97 & 15.81 & 36.96 \\
        +FT       & 63.76 & 60.80 & 55.99 & 15.85 & 36.84 \\
        w/ DMA & 63.48 & 63.00 & \underline{57.51} & \underline{16.13} & \underline{42.07} \\
        w/ CGA       & \underline{63.83} & \underline{64.46} & 56.77 & 15.99 & 39.80 \\
        {\cellcolor{blue!10}}\textbf{Ours} & 
        {\cellcolor{blue!10}}\textbf{64.63} & 
        {\cellcolor{blue!10}}\textbf{67.54} & 
        {\cellcolor{blue!10}}\textbf{57.99} & 
        {\cellcolor{blue!10}}\textbf{16.21} & 
        {\cellcolor{blue!10}}\textbf{42.83} \\
        \bottomrule
        \end{tabular}
}
\end{wraptable}


\subsection{Ablation Studies}
\paragraph{Effectiveness of Proposed Components.} As shown in Table~\ref{tab:ablation}, compared to the baseline and the naive finetuning strategy, our method achieves significant performance gains. Specifically, the DMA module significantly enhances the model's perception of local spatial relationships, boosting the segmentation performance from 36.96 to 42.07. Meanwhile, the CGA module effectively improves accuracy on basic zero-shot domain tasks via implicit semantic alignment. Finally, the full model integrates the strengths of explicit geometric constraints and implicit semantic reconstruction, pushing both fine-grained inference and segmentation capabilities to their peak performance. This verifies the complementarity and necessity of DMA and CGA in enhancing fine-grained understanding.

\begin{wrapfigure}[13]{r}{0.5\textwidth}
\vspace{-4em}
    \centering
    \includegraphics[width=\linewidth]{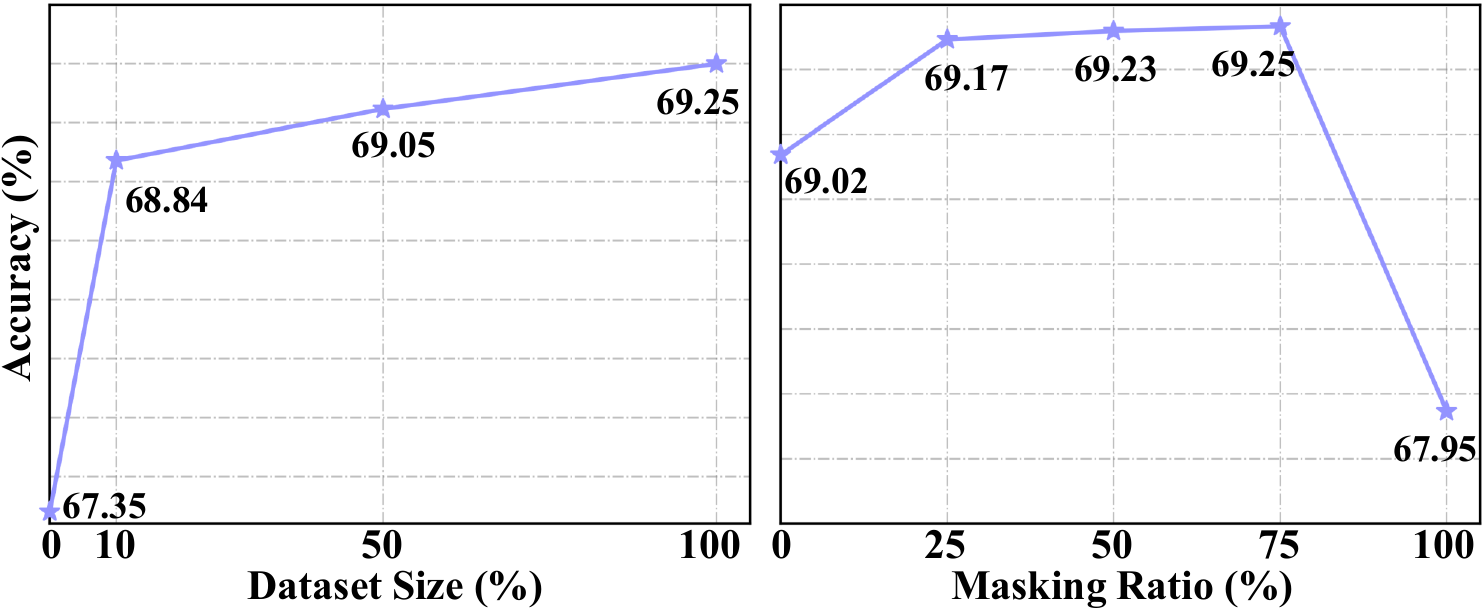}
    
    \caption{Ablation analysis of data scale and masking ratio evaluated on ZS. datasets. Left: Performance scales consistently with increasing data volume. Right: SALM exhibits robustness across masking ratios, with 75\% striking an optimal balance for capturing fine-grained semantics while retaining necessary spatial anchors.}
    \label{fig:size}
\end{wrapfigure}

\paragraph{Impact of Data Scale.} 
We evaluate the scalability of our method by varying the data size. As illustrated in Figure~\ref{fig:size}, we observe consistent performance gains as the data volume increases. This trend highlights that our method is lightweight yet highly scalable, demonstrating the potential to effectively leverage significantly larger data distributions.

\paragraph{Impact of Masking Ratio.} We vary the masking ratio to analyze its impact. Results in Figure~\ref{fig:size} demonstrate that SALM exhibits remarkable robustness across a wide range of ratios. Lower ratios allow the model to rely trivially on local interpolation, while complete masking fails to preserve geometric correspondence due to the lack of spatial anchors. A masking ratio of 75\% strikes an optimal balance, forcing the model to effectively capture fine-grained semantics.


\subsection{Visualization Analysis}
We visualize the PCA-based feature maps of different encoders in Figure~\ref{fig:teaser}. The original CLIP features exhibit coarse and scattered activation patterns with blurred boundaries, confirming its limitation in capturing local geometric details. While DINOv2 demonstrates superior structural coherence, previous alignment methods like un$^2$CLIP and KUEA struggle to fully reconstruct this spatial precision, often retaining significant background noise or ambiguous edges. In contrast, our method generates feature maps with remarkably clear semantic layouts, effectively distinguishing foreground objects from the background.

\section{Conclusions}
In this paper, we propose SALM, a novel unsupervised alignment framework that effectively reconciles the tension between the robust global semantics of CLIP and the fine-grained structural priors of vision-centric models. Unlike prior kernel-based or distillation approaches, SALM synergizes the explicit geometric constraints of Dual-Matrix Alignment with the implicit reconstruction of latent masked modeling. This dual-pathway mechanism successfully injects intricate geometric details into the CLIP embedding space without compromising its pre-trained manifold integrity. Extensive experiments validate that SALM yields consistent gains in zero-shot tasks while achieving superior performance on dense prediction benchmarks. Furthermore, our aligned encoder significantly empowers MLLMs with versatile fine-grained capabilities, ranging from precise object counting to complex spatial reasoning, in a cost-effective manner. We hope this work establishes an efficient paradigm for evolving foundational vision encoders towards more unified multimodal understanding.

\begin{ack}
This work was supported by National Natural Science Foundation of China (NSFC) 62672187 and 
GuangDong Basic and Applied Basic Research Foundation 2025B1515120037.
\end{ack}

{
\small
\bibliographystyle{plain}
\bibliography{example_paper}
}


\appendix
\onecolumn
\section*{\centering Unlocking Fine-Grained Perception in CLIP via Structurally-Aware Latent Masked Modeling (Appendix)}

\hypersetup{linkbordercolor=black,linkcolor=black}

\setlength{\cftbeforesecskip}{0.5em}
\cftsetindents{section}{0em}{1.8em}
\cftsetindents{subsection}{1em}{2.5em}

\etoctoccontentsline{part}{Appendix}
\localtableofcontents
\hypersetup{linkbordercolor=red,linkcolor=red}

\section{More Implementation Details}

\subsection{Regularization}
To prevent catastrophic forgetting and maintain a healthy feature space distribution, we introduce two regularization terms:
\begin{itemize}
    \item \textbf{Reference Regularization~\cite{schlarmann2024robustclip,gong2025kernel}:} We freeze the weights of the original CLIP model to serve as a reference. The global features of the current model are constrained using MSE losses to ensure semantic consistency.
    \item \textbf{Uniformity Loss~\cite{zhang2022mask}:} Applied to the CLS token, this loss prevents feature collapse and maintains the uniformity of the feature distribution.
\end{itemize}

\subsection{Three-Stage Curriculum Strategy}
To effectively balance the difficulty between structural learning and feature reconstruction, we design a dynamic three-stage curriculum strategy that adjusts the weights of $\mathcal{L}_{\text{dma}}$ and $\mathcal{L}_{\text{mra}}$ (denoted as $\gamma$ and $\beta$) over the course of the training. 
In the first stage, Structural Warm-up (1 epoch), we exclusively enable DMA to prioritize learning DINO's macroscopic spatial geometric structures without interference from specific feature values. 
Subsequently, in the Progressive Transition stage (2 epochs), we employ linear interpolation to gradually decay $\gamma$ from $1.0$ to $0.1$ while simultaneously increasing $\beta$ from $0.0$ to $1.0$. This phase smoothly introduces the more challenging reconstruction task while gradually relaxing the strong global structural constraints. 
Finally, during the Fine-grained Alignment stage (1 epoch), the training focuses on precise feature reconstruction dominated by MRA ($\beta = 1.0$), with DMA kept at a minimal weight ($\gamma = 0.1$) as an auxiliary constraint to achieve high-fidelity alignment with DINO features.
\textbf{Notably, we emphasize that our core alignment framework is inherently robust}; even without this curriculum strategy, the model achieves significant performance gains over the baseline CLIP across all evaluated tasks. As demonstrated in our ablation studies (see Section ~\ref {3stage}), while the three-stage schedule is not a prerequisite for convergence, it serves as a highly effective optimization heuristic that further stabilizes the training trajectory and unlocks the model's peak fine-grained potential. It is worth noting that for the SALM-Self variant, since the reconstruction loss is removed, we simply apply constant loss weights throughout the entire training process.

\subsection{Experimental Setup}
We trained our model on the ImageNet-1K dataset using 4 NVIDIA RTX 4090 24GB GPUs. The detailed hyperparameter configurations are listed in Table \ref{tab:hyperparams}.

\begin{table}[h]
    \centering
    \caption{Detailed Hyperparameter Configurations.}
    \label{tab:hyperparams}
    \begin{tabular}{l|l}
    \toprule
    \textbf{Hyperparameter} & \textbf{Value / Description} \\
    \midrule
    Dataset & ImageNet-1K (Train split) \\
    Default Input Resolution & $336 \times 336$ \\
    Global Batch Size & 16  \\
    Cross-Guided Adapter Layers & 4 \\
    Optimizer & AdamW \\
    Learning Rate & $1 \times 10^{-5}$ \\
    Weight Decay & $0.05$ \\
    LR Scheduler & Cosine Annealing \\
    Total Training Epochs & 4 \\
    Reg Loss Weight & 5.0 \\
    Uni Loss Weight & 0.01 \\
    \bottomrule
    \end{tabular}
\end{table}
\section{More Experimental Results}
\label{more_details}




\subsection{Generalization across Architectures and Resolutions}

\begin{table*}[!t]
    \centering
    \caption{Generalization and flexibility analysis. We report the performance across various model combinations. The columns \textit{Input Res.} and \textit{Feat. Map} denotes the input resolution and the extracted feature map size, respectively.}
    \label{tab:appendix_generalization}
    
    \resizebox{\textwidth}{!}{
        \begin{tabular}{l | c c | c | cccccccccccc}
            \toprule
            
            \multirow{6}{*}{\textbf{Method}} & 
            \multicolumn{2}{c|}{\textbf{Settings}} & 
            \multirow{6}{*}{\rotatebox{270}{ImageNet}} & 
            \multicolumn{12}{c}{\textbf{Zero-Shot Performance}} \\
            
            \cmidrule(lr){2-3} \cmidrule(lr){5-16}
            
            & \rotatebox{270}{\textbf{Input Res.}} & \rotatebox{270}{\textbf{Feat. Map}} & 
            & \rotatebox{270}{CIFAR-10} & \rotatebox{270}{CIFAR-100} & \rotatebox{270}{Pets} & 
            \rotatebox{270}{Caltech-101} & \rotatebox{270}{RESISC45} & \rotatebox{270}{PCam} & 
            \rotatebox{270}{DTD} & \rotatebox{270}{EuroSAT} & 
            \rotatebox{270}{ImageNet-O} & \rotatebox{270}{FER2013} & \rotatebox{270}{ImageNetV2} & 
            \rotatebox{270}{\textbf{Average}} \\
            \midrule
            
            CLIP (ViT-L-14) & 224$^2$ & $16^2$ & 
            75.53 & 95.59 & 75.77 & 93.18 & 83.27 & 63.36 & 51.98 & 55.42 & 62.53 & 32.25 & 49.86 & 69.84 & 66.64 \\
            
            CLIP (ViT-L-14-336) & 336$^2$ & $24^2$ & 
            76.55 & 94.92 & 74.33 & 93.68 & 83.43 & 63.78 & 60.72 & 55.69 & 61.50& 32.75 & 49.11 & 70.89 & 67.35 \\

            SigLIP & 224$^2$ & $16^2$ & 
            82.02 & 96.74 & 84.12 & 95.33 & 86.05 & 69.68 & 50.58 & 71.01 & 62.98& 29.55 & 49.83 & 76.06 & 70.18 \\
            \midrule

            CLIP (L-14) + MAE & 224$^2$ / 224$^2$ & $\mathbf{16^2 / 14^2}$ & 
            76.13 & 95.63 & 77.29 & 93.51 & 84.29 & 63.48 & 50.4 & 54.95 & 61.85 & 36.6 & 50.68 & 70.23 & \textbf{67.17} \\

            CLIP (L-14) + DINOv2 & 224$^2$ / 224$^2$ & $16^2$ / $16^2$ & 
            76.00 & 96.42 & 77.19 & 93.62 & 84.38 & 64.40 & 53.13 & 56.06 & 62.59 & 36.75 & 50.15 & 69.86 & \textbf{67.69} \\

            CLIP + DINOv2 & 336$^2$ / 336$^2$ & $24^2$ / $24^2$ & 
            77.21 & 96.38 & 77.87 & 94.09 & 84.04 & 64.63 & 67.54 & 56.27 & 62.19 & 37.05 & 50.11 & 71.53 & \textbf{69.25} \\

            SigLIP + DINOv2 & 224$^2$ / 224$^2$ & $16^2$ / $16^2$ & 
            82.38 & 97.63 & 84.55 & 95.67 & 86.26 & 69.13 & 51.04 & 71.48 & 63.68 & 36.00 & 50.84 & 76.74 & \textbf{71.18} \\

            CLIP + DINOv3 & 336$^2$ / 336$^2$ & $\mathbf{24^2 / 21^2}$ & 
            77.23 & 96.16 & 77.75 & 93.79 & 83.89 & 64.47 & 67.07 & 55.90 & 60.22 & 37.55 & 52.79 & 71.49 & \textbf{69.19} \\
            
            \bottomrule
        \end{tabular}
    }
\end{table*}



To comprehensively evaluate the generalization and flexibility of our proposed framework, we conducted extensive experiments on various combinations of VLMs and Vision-Centric Models. We aim to verify whether the framework can effectively adapt to diverse model architectures, as well as varying input resolutions and feature map sizes. Specifically, we introduced several representative model variants in our experimental setup: regarding VLMs, in addition to the standard OpenAI CLIP (ViT-L-14 and ViT-L-14-336)~\cite{radford2021learning}, we also tested the architecturally distinct and more powerful Google SigLIP (so400m-Patch14-224)~\cite{zhai2023sigmoid}. For Vision-Centric Models, we utilized MAE (ViT-MAE-Large)~\cite{he2022masked}, DINOv2 (ViT-L/14 with Registers)~\cite{darcet2024vision} and the latest DINOv3 (VIT-L/16-Pretrain-lvd1689m)~\cite{simeoni2025dinov3}.

As shown in Table \ref{tab:appendix_generalization}, different model combinations result in significant discrepancies in input resolutions and feature map sizes. For instance, the input resolution for CLIP is $336^2$ (feature map is $24^2$), whereas for SigLIP it is $224^2$ (feature map size is $16^2$); in the CLIP + DINOv3 combination, the feature map sizes are $24^2$ and $21^2$, respectively, and similarly, in the CLIP + MAE combination, the feature map sizes are $16^2$ and $14^2$. To address the inconsistency between the feature map resolutions of the source and target models, our framework demonstrates exceptional adaptability: we resize the features of the source model via interpolation to match the feature dimensions of the target model, thereby achieving effective feature alignment and fusion. Furthermore, to address discrepancies in feature embedding dimensions between the two encoders, we project the source model's features onto the target dimension via a linear layer.

Experimental results indicate that our method significantly enhances the zero-shot performance of VLMs, regardless of variations in base architecture or resolution. By incorporating MAE as the source model, the average accuracy of the baseline CLIP (ViT-L-14) across 11 datasets improves from 66.64\% to 67.17\%. After introducing DINOv2 as the source model, the average accuracy of CLIP across 11 datasets improved from 67.35\% to 69.25\%; the more advanced SigLIP also saw an improvement from 70.18\% to 71.18\%, demonstrating the universality of our framework across different VLMs. 

Even when replacing the model with DINOv3 and facing the challenge of mismatched feature map sizes, the CLIP + DINOv3 combination still achieved an outstanding performance of 69.19\%, surpassing the original CLIP. Notably, despite the spatial interpolation required by this structural mismatch, DINOv3 exhibits distinct advantages in handling distribution shifts and specialized fine-grained domains. Specifically, it outperforms the DINOv2 counterpart on ImageNet-O ($37.55\%$ \textit{v.s.} $37.05\%$) and FER2013 ($52.79\%$ \textit{v.s.} $50.11\%$). These results compellingly demonstrate the high flexibility and generalization capability of our framework. It not only allows for the seamless replacement of various VLMs and Vision-Centric Models but also effectively resolves resolution discrepancies between models through interpolation strategies, consistently maintaining stable performance gains and leveraging the unique strengths of different vision encoders.

\begin{table*}[!t]
    \centering
    \caption{Ablation analysis of simple feature distillation strategy and three-stage curriculum strategy on zero-shot benchmarks. We report Top-1 accuracy (\%) on ImageNet-1K and 11 other datasets. The best results are highlighted in \textbf{bold}.}
    \label{tab:3stage_cmp}
    \resizebox{\textwidth}{!}{
        \begin{tabular}{l | c | cccc cccc cccc}
            \toprule
            \multirow{7}{*}{\textbf{Method}} & 
            \multirow{7}{*}{\rotatebox{270}{ImageNet}} & 
            \multicolumn{12}{c}{\textbf{Zero-Shot}} \\
            
            \cmidrule(lr){3-14}
            
             & & 
            \rotatebox{270}{CIFAR-10} & \rotatebox{270}{CIFAR-100} & \rotatebox{270}{Pets} & 
            \rotatebox{270}{Caltech-101} & \rotatebox{270}{RESISC45} & \rotatebox{270}{PCam} & 
            \rotatebox{270}{DTD} & \rotatebox{270}{EuroSAT} & 
            \rotatebox{270}{ImageNet-O} & \rotatebox{270}{FER2013} & \rotatebox{270}{ImageNetV2} & 
            \rotatebox{270}{\textbf{Average}} \\ 
            \midrule
            
            CLIP & 76.55 & 94.92 & 74.33 & 93.68 & 83.43 & 63.78 & 60.72 & 55.69 & 61.50 & 32.75 & 49.11 & 70.89 & 67.35 \\

            Simple Distillation & \underline{77.06} & 95.94 & \underline{77.40} & 93.13 & 83.09 & \underline{64.39} & 58.26 & 55.85 & 60.92 & \textbf{37.90} & 49.27 & 71.39 & 67.96 \\

            Ours w/o 3-stage & 77.02 & \underline{96.30} & 77.24 & \underline{93.97} & \textbf{84.07} & 64.10 & \underline{62.89} & \underline{55.95} & \underline{62.16} &  36.65 & \underline{49.81} & \underline{71.50} & \underline{68.60} \\

            {\cellcolor{blue!10}}\textbf{Ours w/ 3-stage} &{\cellcolor{blue!10}{\textbf{77.21}}} &{\cellcolor{blue!10}{\textbf{96.38}}} &{\cellcolor{blue!10}{\textbf{77.87}}} &{\cellcolor{blue!10}{\textbf{94.09}}} &{\cellcolor{blue!10}{\underline{{84.04}}}} &{\cellcolor{blue!10}{\textbf{64.63}}} &{\cellcolor{blue!10}{\textbf{67.54}}} &{\cellcolor{blue!10}{\textbf{56.27}}} &{\cellcolor{blue!10}{\textbf{62.19}}} &{\cellcolor{blue!10}{{\underline{37.05}}}} &{\cellcolor{blue!10}{\textbf{50.11}}} &{\cellcolor{blue!10}{\textbf{71.53}}} &{\cellcolor{blue!10}{\textbf{69.25}}} \\
            
            \bottomrule
        \end{tabular}
    }
\end{table*}
\begin{table}[t]
\centering
\caption{Ablation analysis of simple feature distillation strategy and three-stage curriculum strategy on zero-shot fine-grained understanding and dense prediction. We report Top-1 accuracy (\%) for zero-shot fine-grained understanding tasks, alongside mean IoU (mIoU) for semantic segmentation.}

\label{tab:3_stage_fine_seg}
\resizebox{0.6\linewidth}{!}{
\begin{tabular}{l|ccc|c}
\toprule
\multirow{2.5}{*}{\textbf{Method}} & \multicolumn{3}{c|}{\textbf{ZS. Fine-grained Tasks}} & \multicolumn{1}{c}{\textbf{LP. Seg.}} \\
 & SVHN &\adjustbox{stack=cc}{CLEVR\\Distance}&\adjustbox{stack=cc}{CLEVR\\Counts} &ADE20K\\
\midrule
CLIP      & 55.97 & 15.81 & 20.01 &  36.96 \\
Simple Distillation  & 51.87 & 15.92 & 22.63 & \textbf{43.09} \\
Ours w/o 3-stage & \underline{57.95} & \underline{15.93} & \underline{22.78} & 42.29 \\

{\cellcolor{blue!10}}\textbf{Ours w/ 3-stage} & 
{\cellcolor{blue!10}}\textbf{57.99} & 
{\cellcolor{blue!10}}\textbf{16.21} & 
{\cellcolor{blue!10}}\textbf{22.81} & 
{\cellcolor{blue!10}}\underline{42.83} \\
\bottomrule
\end{tabular}
}
\vspace{-3mm} 
\end{table}
\subsection{Compare with Simple Feature Distillation}
To further verify the necessity of our proposed dual-matrix alignment and masked reconstruction alignment modules, we compared SALM with a naive feature distillation baseline. In this baseline, the source CLIP features are first mapped via a linear projection layer to match the target dimension. We then directly minimize the Mean Squared Error (MSE) between the projected CLIP features and the target DINOv2 features at both the CLS token and patch token levels, while incorporating a regularization term $\mathcal{L}_{\text{reg}}$ to preserve the original semantic space. The objective is defined as:$$\mathcal{L} = \|h_{\omega}(\mathbf{Z}^{\text{src}}) - \mathbf{Z}^{\text{ref}}\|_2^2 + \alpha \mathcal{L}_{\text{reg}},$$where $h_{\omega}$ denotes the linear projection layer.

As shown in Table~\ref{tab:3stage_cmp} and \ref{tab:3_stage_fine_seg}, experimental results reveal the limitations of relying solely on point-wise feature alignment. While simple distillation achieves high performance on the ADE20K semantic segmentation task, indicating that strong constraints can inject DINO information, this comes at the cost of the model's general semantic representation capability. This trade-off is particularly evident in the PCam and SVHN task, where the accuracy of simple distillation drops significantly from the original CLIP's 55.97\% to 51.87\%, suggesting that overly rigid feature approximation disrupts CLIP's pre-trained semantic manifold and leads to catastrophic forgetting. In contrast, SALM maintains a high level of performance on dense prediction tasks comparable to simple distillation, while significantly outperforming simple distillation in fine-grained understanding and overall zero-shot classification. This demonstrates that SALM's dual-matrix alignment and masked reconstruction mechanisms are not redundant; rather than simply replicating the teacher model's values, they learn its structure and contextual relationships to inject fine-grained details while effectively protecting and enhancing the original multimodal semantic space, thereby achieving an optimal balance between accuracy and robustness.
\begin{figure}
    \centering
    \includegraphics[width=0.8\linewidth]{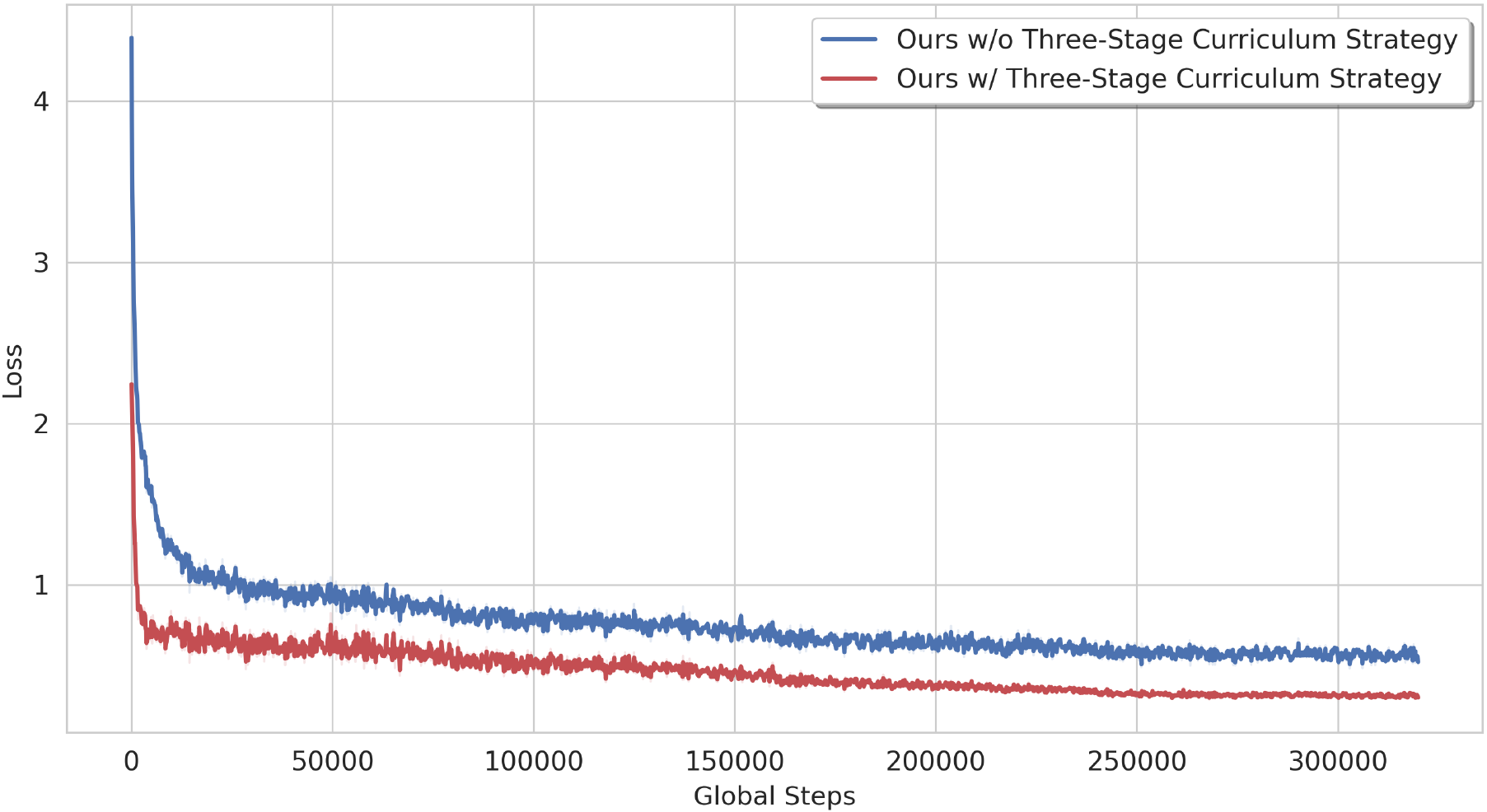}
    \caption{Ablation analysis of the training schedule. We compare the weighted total objective loss curves of models trained with and without the three-stage curriculum strategy. Both settings successfully converge, showing that the model can still train and converge stably when the curriculum strategy is removed.}
    \label{fig:training_loss}
\end{figure}
\subsection{Impact of Three-Stage Curriculum Strategy}
\label{3stage}
Our default training pipeline employs a three-stage curriculum strategy (Structural Warm-up $\rightarrow$ Progressive Transition $\rightarrow$ Fine-grained Alignment) to stabilize the optimization process. To assess the sensitivity of our framework to this training schedule, we conducted an ablation study where the model was trained without the curriculum strategy, utilizing constant loss weights ($\alpha, \beta, \gamma$) from the beginning. Experimental results in Figure~\ref{fig:training_loss} demonstrate that SALM is robust to the removal of the curriculum strategy, successfully converging without divergence. As shown in Table~\ref{tab:3stage_cmp} and \ref{tab:3_stage_fine_seg}, however, we observe a slight performance degradation in accuracy compared to the full three-stage training. This indicates that while the curriculum strategy is not strictly mandatory for convergence, it serves as an effective optimization heuristic. Prioritizing coarse-grained structural constraints before introducing fine-grained reconstruction objectives helps the model navigate the optimization landscape more smoothly, thereby unlocking the model's peak potential in capturing intricate visual details.

\subsection{Decoupled Ablation Analysis of Dual-Matrix Alignment}
\begin{table}[t]
\centering
\caption{Ablation analysis of the Dual-Matrix Alignment (DMA) strategy on various downstream tasks. We decouple the effects of the Spatial Relation Matrix ($K$) and the Energy Difference Matrix ($D$).}

\label{tab:dma_decoupled}
\resizebox{0.5\linewidth}{!}{
\begin{tabular}{l|ccc|c}
\toprule
\multirow{2.5}{*}{\textbf{Method}} & \multicolumn{3}{c|}{\textbf{ZS. Fine-grained Tasks}} & \multicolumn{1}{c}{\textbf{LP. Seg.}} \\
 & SVHN &\adjustbox{stack=cc}{CLEVR\\Distance}&\adjustbox{stack=cc}{CLEVR\\Counts} &ADE20K\\
\midrule
w/o DMA      & 56.77 & 15.99 & 20.99 & 39.80 \\
w/ $K$  & 56.73 & \underline{16.05} & 22.17 & \underline{40.98} \\
w/ $D$ & \underline{56.87} & 16.01 & \underline{22.39} & 39.97 \\

{\cellcolor{blue!10}}\textbf{w/ DMA} & 
{\cellcolor{blue!10}}\textbf{57.99} & 
{\cellcolor{blue!10}}\textbf{16.21} & 
{\cellcolor{blue!10}}\textbf{22.81} & 
{\cellcolor{blue!10}}\textbf{42.83} \\
\bottomrule
\end{tabular}
}
\vspace{-3mm} 
\end{table}
To further investigate the individual contributions of the Spatial Relation Matrix ($K$) and the Energy Difference Matrix ($D$) within the Dual-Matrix Alignment (DMA) strategy, we conducted additional decoupled ablation experiments across various downstream tasks. As shown in Table \ref{tab:dma_decoupled}, the experimental results thoroughly validate the complementarity of these two matrices in capturing different dimensions of geometric priors.

In dense prediction tasks that impose stringent requirements on local coherence (semantic segmentation on ADE20K), the model incorporating solely the Spatial Relation Matrix (w/ $K$) achieved an mIoU of 40.98, significantly outperforming the variant relying exclusively on the Energy Difference Matrix (w/ $D$), which obtained 39.97. This indicates that by constraining the cosine similarity between features, the $K$ matrix successfully injects geometric priors essential for understanding local topological structures and delineating precise object boundaries. The advantage of the $D$ matrix becomes more pronounced in the counting task, scoring 22.39 compared to 22.17 for the $K$ matrix. The core challenge in counting tasks lies in isolating spatially connected or visually similar objects into distinct individual entities. By characterizing the variance in feature norms, the $D$ matrix accurately captures the relative saliency ranking of local regions, thereby effectively facilitating the separation of discrete instances.

Finally, a striking synergistic effect is observed in the complex number recognition task (SVHN), which is characterized by extremely noisy backgrounds. Relying solely on either the $K$ matrix or the $D$ matrix resulted in a performance bottleneck, achieving results comparable to or slightly lower than the baseline model with only implicit reconstruction (w/ CGA, 56.77). However, the combination of both matrices (w/ DMA) yields a substantial performance leap to 57.99. This comprehensively demonstrates that in highly distractive environments, the model intrinsically requires the $D$ matrix to suppress background noise and focus on salient targets, while simultaneously depending on the $K$ matrix to discern fine-grained geometric topologies. 

\subsection{Reference Layer Selection for SALM-Self}
SALM-Self utilizes self-generated shallow features as the reference to provide local structural guidance without relying on external vision models. The choice of reference layer is important, as shallow layers contain richer local details but may lack semantic abstraction, while deeper layers gradually aggregate global semantics and lose fine-grained spatial information. Therefore, we investigate different CLIP layers as the self-reference feature to analyze the trade-off between local structure preservation and semantic consistency.

As shown in Table~\ref{tab:salmself_layer}, we evaluate different layers of CLIP as the reference feature. The 6th layer achieves the best overall performance, obtaining 77.13\% Top-1 accuracy on ImageNet-1K and 68.60\% average performance across evaluation benchmarks. Meanwhile, other nearby layers achieve comparable results, indicating that SALM-Self is not highly sensitive to the exact layer selection. These results suggest that intermediate layers provide an effective balance between retaining local visual structures and maintaining semantic consistency.

\begin{table}[t]
\centering
\caption{Ablation study on the reference layer selection for SALM-Self.}
\label{tab:salmself_layer}
\resizebox{0.45\textwidth}{!}{
\begin{tabular}{l|cc}
\toprule
\textbf{Reference Layer} & \textbf{ImageNet-1K} & \textbf{Avg.} \\
\midrule
2  & 76.90 & 68.57 \\
4  & 77.06 & 68.35 \\
6  & \textbf{77.13} & \textbf{68.60} \\
8  & 77.07 & 68.43 \\
10 & 76.95 & 68.58 \\
\bottomrule
\end{tabular}
}
\vspace{-3mm}
\end{table}

\subsection{Evaluation on Open-Vocabulary Segmentation}
Although SALM is not specifically designed for open-vocabulary segmentation, we evaluate it in this setting to examine whether the enhanced local representations remain compatible with CLIP's original text space. Specifically, we compare SALM with CLIP and DeCLIP under the open-vocabulary segmentation protocol. As shown in Table~\ref{tab:ovss}, SALM consistently improves over the original CLIP across all evaluated datasets, indicating that enhancing local visual structures does not compromise the compatibility between CLIP visual features and text embeddings. However, SALM remains inferior to DeCLIP, which achieves substantially better performance by explicitly modeling content/context features and optimizing dense alignment for open-vocabulary perception. We emphasize that this result does not indicate that SALM explicitly improves patch-text alignment, as SALM does not introduce region-level or patch-level text supervision. Instead, the improvement mainly comes from enhancing the intrinsic spatial structure and semantic quality of CLIP representations. 

\begin{table}[t]
\centering
\caption{Open-vocabulary segmentation results. We report mIoU (\%) on five segmentation benchmarks.}
\label{tab:ovss}
\resizebox{0.7\textwidth}{!}{
\begin{tabular}{l|ccccc}
\toprule
\textbf{Method} & VOC2012 & Context & COCO-Stuff & Cityscapes & ADE20K \\
\midrule
CLIP & 14.79 & 4.06 & 2.04 & 1.28 & 1.43 \\
Simple Distillation & 2.07 & 0.29 & 0.05 & 0.16 & 0.02 \\
DeCLIP & \textbf{85.16} & \textbf{39.37} & \textbf{28.75} & \textbf{33.13} & \textbf{21.92} \\

{\cellcolor{blue!10}}\textbf{SALM (Ours)} &
{\cellcolor{blue!10}}\underline{17.03} &
{\cellcolor{blue!10}}\underline{7.75} &
{\cellcolor{blue!10}}\underline{4.22} &
{\cellcolor{blue!10}}\underline{3.08} &
{\cellcolor{blue!10}}\underline{3.34} \\

\bottomrule
\end{tabular}
}
\vspace{-3mm}
\end{table}

\subsection{Evaluation on MMVP VLM Benchmark}
To further evaluate the model's capability in capturing fine-grained visual details and spatial relationships, we conducted evaluations on the MMVP(Multimodal Visual Patterns)-VLM benchmark~\cite{tong2024eyes}, which is specifically optimized for VLMs by simplifying language descriptions to isolate visual discrimination capabilities. 
It is organized into 9 distinct visual patterns, with each pattern containing 15 pairs of zero-shot questions. 
Unlike standard classification metrics, MMVP-VLM specifically challenges the VLM's ability to distinguish between visually similar images that require precise structural understanding, effectively probing the ``visual blind spots'' often found in CLIP-like models.

Table \ref{tab:mmvp_vlm} presents the quantitative comparison on the MMVP-VLM benchmark. 
Our method achieves an average accuracy of 25.9\%, significantly outperforming both the CLIP baseline (20.0\%) and the competitive method KUEA (21.5\%). 
This substantial improvement (+5.9\% over CLIP) validates that distilling fine-grained features effectively mitigates the ``visual blind spots'' inherent in CLIP. Specifically, we observe remarkable gains in categories requiring precise spatial and attribute discrimination. Most notably, in the \textit{Orientation and Direction} category (\textbf{\faCompass}), where the baseline completely fails (0.0\%), our method successfully recovers valid predictions, achieving an accuracy of 13.3\%. Furthermore, our approach demonstrates superior capability in capturing \textit{Color and Appearance} (\textbf{\faPalette}), improving accuracy from 20.0\% to 53.3\%, and doubles the performance in \textit{Positional and Relational Context} (\textbf{\faMapPin}) from 6.7\% to 13.3\%. 
These results confirm that the structural knowledge learned through our proposed Dual Matrix Alignment and Masked Reconstruction tasks significantly enhances the model's fine-grained visual reasoning capabilities.
\begin{table*}[!t]
    \centering
    \caption{Performance of CLIP on MMVP-VLM benchmark. Symbols for visual patterns are inherited:
    \textbf{\faCompass}: Orientation and Direction, \textbf{\faSearch}: Presence of Specific Features, \textbf{\faSync}: State and Condition, \textbf{\faSortNumericUp}: Quantity and Count, \textbf{\faMapPin}: Positional and Relational Context, \textbf{\faPalette}: Color and Appearance, \textbf{\faCogs}: Structural and Physical Characteristics, \textbf{\faFont}: Texts, \textbf{\faCamera}: Viewpoint and Perspective.}
    \label{tab:mmvp_vlm}
    \resizebox{0.7\textwidth}{!}
    {
        \begin{tabular}{l | ccccccccc | c}
            \toprule
            \textbf{Method} & 
             \faCompass & \faSearch & \faSync & \faSortNumericUp & \faMapPin & \faPalette & \faCogs & \faFont & \faCamera & 
            \textbf{Average} \\
            \midrule
            
            CLIP & 0.0 & 20.0 & 40.0 & \textbf{20.0} & 6.7 & 20.0 & \textbf{33.3} & 6.7 & \textbf{33.3} & 20.0 \\

            KUEA & 6.6 & \textbf{26.7} & 40.0 & 13.3 & 6.7 & 40.0 & 26.6 & \textbf{13.3} & 20.0  & 21.5 \\
            
            {\cellcolor{blue!10}}\textbf{Ours} &
            {\cellcolor{blue!10}}\textbf{13.3} &
            {\cellcolor{blue!10}}20.0 &
            {\cellcolor{blue!10}}\textbf{46.7} &
            {\cellcolor{blue!10}}13.3 &
            {\cellcolor{blue!10}}\textbf{13.3} &
            {\cellcolor{blue!10}}\textbf{53.3} &
            {\cellcolor{blue!10}}\textbf{33.3} &
            {\cellcolor{blue!10}}\textbf{13.3} &
            {\cellcolor{blue!10}}26.7 &
            {\cellcolor{blue!10}}\textbf{25.9} \\
            
            \bottomrule
        \end{tabular}
    }
\end{table*}

\subsection{Evaluation on Zero-shot Image-Text Retrieval}




\begin{table*}[!t]
\centering
\caption{Zero-shot image-text retrieval results on Flickr30K and MSCOCO benchmarks. We report Recall@K (K=1, 5, 10) for both Image-to-Text and Text-to-Image retrieval tasks.}
\label{retrieval}

\footnotesize
\resizebox{0.9\textwidth}{!}{

\begin{tabular}{l|ccc|ccc|ccc|ccc}
\toprule

\multirow{3}{*}{\adjustbox{stack=cc}{\textbf{Method}}}&\multicolumn{6}{c|}{\textbf{Image-to-Text Retrieval}}&\multicolumn{6}{c}{\textbf{Text-to-Image Retrieval}}\\
&\multicolumn{3}{c|}{Flickr30K}&\multicolumn{3}{c|}{MSCOCO}&\multicolumn{3}{c|}{Flickr30K}&\multicolumn{3}{c}{MSCOCO}\\
&R@1&R@5&R@10&R@1&R@5&R@10&R@1&R@5&R@10&R@1&R@5&R@10\\
\midrule

CLIP& \textbf{87.30} & 98.29 & \textbf{99.30} & 57.92 & 81.20 & 87.86 & 67.34 & 89.02 & 93.28 & 37.07 & 61.64 & 71.51 \\
KUEA& \textbf{87.30} & 98.10 & \textbf{99.30} & 58.26 & 81.24 & 88.27 & 68.70 & \textbf{89.80} & 93.83 & 38.05 & 62.64 & 72.60 \\
{\cellcolor{blue!10}}\textbf{Ours} & {\cellcolor{blue!10}{86.70}} & {\cellcolor{blue!10}{\textbf{98.40}}} & {\cellcolor{blue!10}{99.10}}&{\cellcolor{blue!10}{\textbf{59.12}}} & {\cellcolor{blue!10}{\textbf{82.08}}} & {\cellcolor{blue!10}{\textbf{88.36}}} & {\cellcolor{blue!10}{\textbf{68.88}}}  & {\cellcolor{blue!10}{\textbf{89.80}}} & {\cellcolor{blue!10}{\textbf{94.20}}} & {\cellcolor{blue!10}{\textbf{38.54}}} & {\cellcolor{blue!10}{\textbf{63.25}}} & {\cellcolor{blue!10}{\textbf{73.10}}} \\
\bottomrule
\end{tabular}}
\vspace{-1mm}
\end{table*}

We further conducted zero-shot image-text retrieval experiments to verify the fine-grained semantic alignment between visual and textual representations. This task evaluates the quality of the feature space directly, ensuring that semantically similar images and texts are projected closely. We utilized two standard benchmarks for this assessment: Flickr30K~\cite{young2014image} and MSCOCO~\cite{chen2015microsoft}. As reported in Table \ref{retrieval}, our method achieves superior or competitive performance compared to CLIP and KUEA across both Image-to-Text and Text-to-Image retrieval tasks. Specifically, our model demonstrates robust improvements on the MSCOCO dataset, confirming that the learned representations maintain high fidelity even without task-specific fine-tuning.

\subsection{Visualization of Global Semantic Features.}
\begin{figure}
    \centering
    \includegraphics[width=\linewidth]{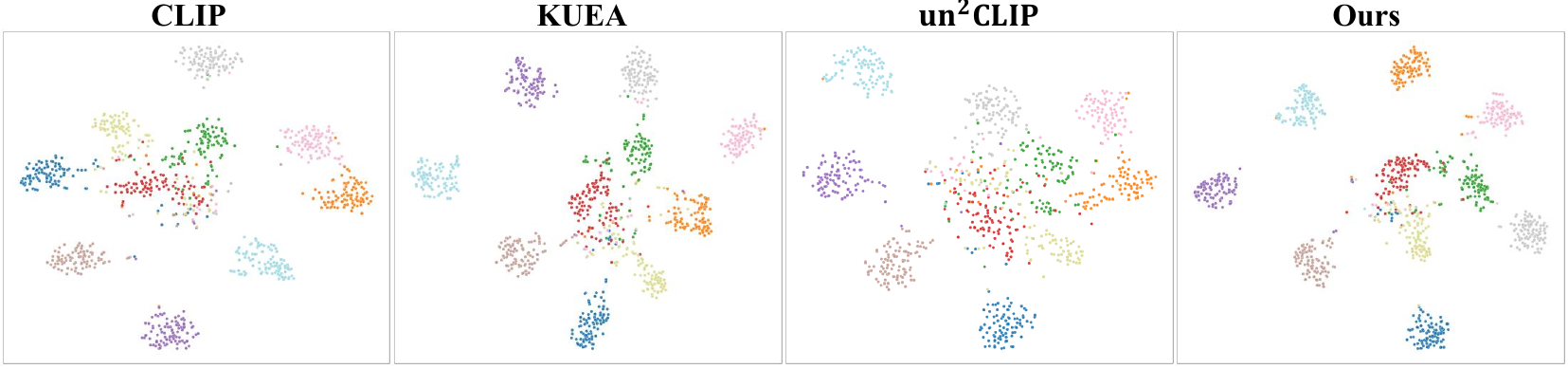}
    \caption{The t-SNE visualization of global semantic features. Compared to other methods, our approach generates more compact and well-separated clusters, demonstrating superior capability in learning discriminative global representations.}
    \label{fig:tsne}
\end{figure}

To further evaluate the quality of the learned representation space, we visualize the distribution of global semantic features using t-SNE in Figure~\ref{fig:tsne}. As illustrated, the embedding clusters of the original CLIP and recent alignment methods appear relatively dispersed, with ambiguous boundaries between adjacent semantic categories. This looseness suggests a latent instability in distinguishing semantically similar samples. In sharp contrast, our method produces a significantly more structured feature space, characterized by high intra-class compactness and clear inter-class separability. This evidence confirms that SALM's alignment strategy effectively integrates fine-grained geometric priors, and it refines the global semantic distribution, resulting in more robust and discriminative representations.

\subsection{More PCA Visualizations}
\begin{figure}
    \centering
    \includegraphics[width=\linewidth]{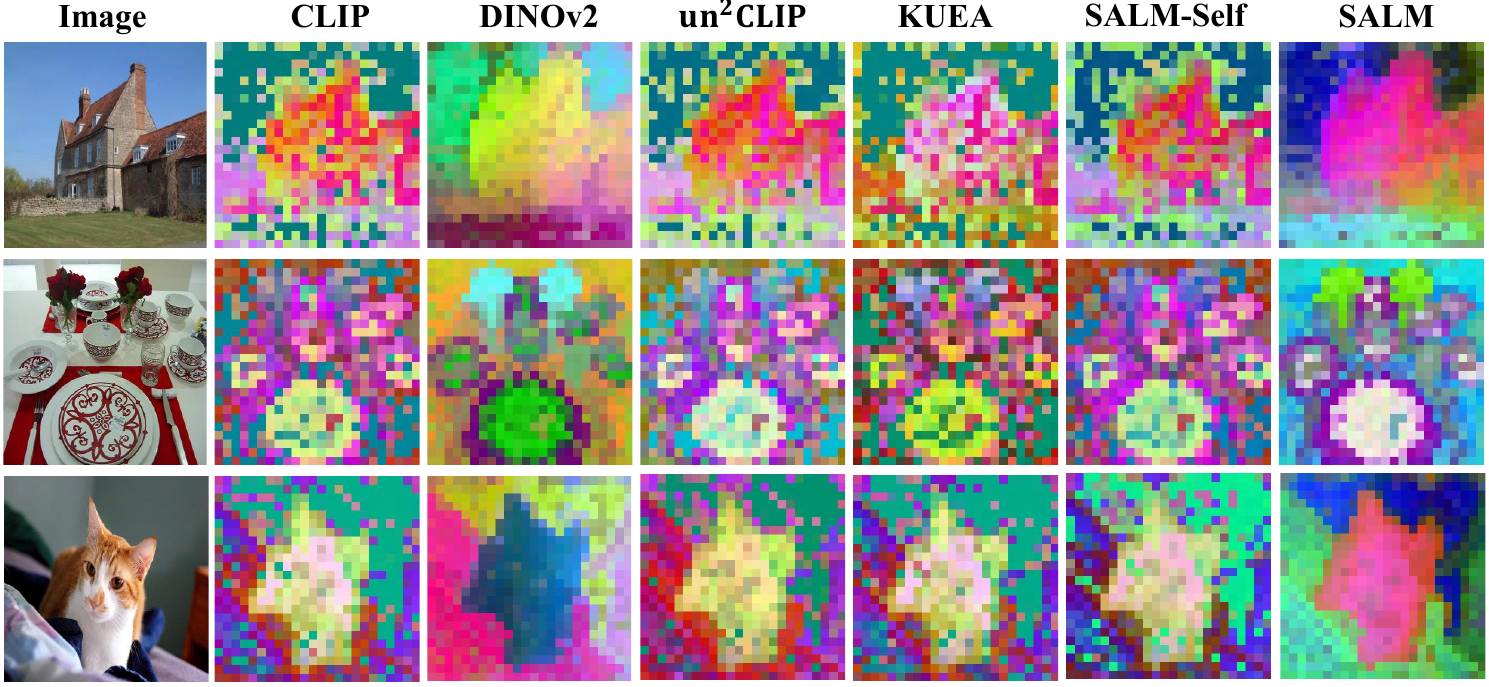}
    \caption{PCA-based visualizations. Unlike other enhancement methods, our approach excels at capturing fine-grained semantic nuances and preserving discriminative local details.}
    \label{fig:pca2}
\end{figure}

As shown in Figure \ref{fig:pca2}, we present additional PCA visualization results to further evaluate the performance of different models at the feature manifold level. It can be observed that the feature distribution of vanilla CLIP exhibits fragmented, blocky artifacts accompanied by substantial background noise, reflecting its inherent limitations in capturing fine-grained local structures. Although existing visual enhancement methods (un$^2$CLIP and KUEA) show marginal improvements, they still suffer from noticeable background interference and ambiguous semantic boundaries. In contrast, our proposed SALM-Self method exhibits remarkable noise reduction capabilities and clear semantic layouts under zero external model dependency. Furthermore, the complete SALM model, which incorporates external priors, achieves structural coherence most akin to DINOv2, accomplishing not only highly pure foreground-background separation but also exceptionally sharp local semantic boundaries, comprehensively validating the remarkable effectiveness of our proposed alignment framework in reshaping high-quality, fine-grained features.

\section{Limitations}
While SALM effectively bridges the gap between global semantics and local structures, it possesses certain inherent limitations. First, the alignment performance is fundamentally capped by the representation quality and granularity of the vision-centric models used as references. Although SALM is model-agnostic and generalizes well across various teachers (\textit{e.g.}, DINOv2/v3, MAE), any structural noise or lack of specific semantic details in the target model's latent space will inevitably be inherited during the distillation process. In scenarios where the target model fails to capture essential geometric priors, the enhancement of CLIP's fine-grained perception may reach a "representation ceiling."
Second, the efficacy of our framework is constrained by the intrinsic architectural capacity and pre-trained stability of the VLM backbone. While we unlock fine-grained perception capabilities within CLIP's features, the fixed parameter count and the necessity of preserving the original multimodal semantic manifold impose a trade-off. Over-injecting dense structural information might eventually risk saturating the representation space, potentially causing a minor interference with the model's global reasoning in extreme edge cases. 
Finally, although our SALM-Self paradigm mitigates the dependency on external experts, it still operates within the boundaries of the VLM's existing pre-trained knowledge. Future research could investigate more advanced feature decoupling mechanisms, allowing the VLM to autonomously acquire out-of-distribution geometric priors without risking global semantic collapse.

\section{Broader Impact}
This paper presents SALM, a framework designed to enhance the fine-grained visual perception of Vision-Language Models. By bridging the gap between global semantics and local structure, our approach has the potential to significantly advance real-world applications where precision is paramount. For instance, in assistive technologies, the improved capability to perform spatial reasoning and accurate object counting can lead to more reliable navigation aids for the visually impaired. Similarly, in embodied AI and autonomous systems, robust topological understanding is essential for safer interaction with complex environments, helping to mitigate perceptual hallucinations in safety-critical scenarios. 
While our method improves technical accuracy, it relies on pre-trained foundation models and may inherit their underlying biases. Furthermore, the heightened capacity for fine-grained visual recognition could potentially be misused for invasive surveillance, unauthorized tracking, or extracting sensitive personal attributes from images.  We encourage the community to deploy these enhanced capabilities responsibly, particularly regarding user privacy and fairness.



\end{document}